\documentclass[10pt,twocolumn,letterpaper]{article}

\usepackage{wacv}
\usepackage{times}
\usepackage{epsfig}
\usepackage{graphicx}
\usepackage{amsmath}
\usepackage{amssymb}
\usepackage{microtype}
\usepackage{graphicx}
\usepackage{booktabs} 
\usepackage{soul}
\usepackage{url}
\usepackage[pagebackref=true,breaklinks=true,letterpaper=true,colorlinks,bookmarks=false]{hyperref}
\usepackage[utf8]{inputenc}
\usepackage[small]{caption}
\usepackage{booktabs}
\usepackage{bclogo}
\usepackage{enumitem}
\usepackage{float}
\usepackage{colortbl}
\usepackage{placeins}
\usepackage{subcaption}
\usepackage{multirow}
\usepackage{pifont}%
\usepackage{gensymb}
\usepackage{glossaries}
\usepackage{comment}

\usepackage{ifthen}
\newboolean{ack}
\newboolean{coordconv}
\newboolean{github}
\setboolean{ack}{true} 
\setboolean{github}{true}
\setboolean{coordconv}{true}

\def\wacvPaperID{658} 

\wacvfinalcopy 

\ifwacvfinal
\fi

\ifwacvfinal
\usepackage[breaklinks=true,bookmarks=false]{hyperref}
\else
\usepackage[pagebackref=true,breaklinks=true,colorlinks,bookmarks=false]{hyperref}
\fi

\ifwacvfinal
\fi

\newglossaryentry{Up Wt}
{
    name=Up Wt,
    description={Group Upweighting}
}
\newglossaryentry{GDRO}
{
    name=GDRO,
    description={Group DRO}
}
\newglossaryentry{RUBi}
{
    name=RUBi,
    description={Reducing Unimodal Biases}
}
\newglossaryentry{LNL}
{
    name=LNL,
    description={Learning Not to Learn}
}
\newglossaryentry{IRMv1}
{
    name=IRMv1,
    description={Invariant Risk Minimization}
}
\newglossaryentry{LFF}
{
    name=LFF,
    description={Learning From Failure}
}
\newglossaryentry{SD}
{
    name=SD,
    description={Spectral Decoupling}
}
\newglossaryentry{BiasedMNIST}
{
    name=Biased MNISTv1,
    description={Biased MNIST}
}
\newglossaryentry{ColoredMNIST}
{
    name=Colored MNIST,
    description={Colored MNIST}
}
\newglossaryentry{conventional}
{
    name=StdM,
    description={Standard Model}
}
\begin{document}
\newcommand{\ck}[1]{\textcolor{red}{CK: #1}}
\newcommand{\warning}[1]{\textcolor{red}{#1}}
\newcommand{\rs}[1]{ \textcolor{blue}{RS: #1}}
\newcommand{\kk}[1]{ \textcolor{magenta}{KK: #1}}
\newcommand{\TOREVIEW}[1]{ \textcolor{red}{TOREVIEW: #1}\textcolor{black}}
\newcommand{\badcolor}[1]{\cellcolor[HTML]{ffd6cc}}
\newcommand{\badcolortwo}[1]{\cellcolor[HTML]{ffad99}}
\newcommand{\goodcolor}[1]{\cellcolor[HTML]{e6eeff}}
\newcommand{\goodcolortwo}[1]{\cellcolor[HTML]{b3ccff}}
\newcommand{\goodcolorthree}[1]{\cellcolor[HTML]{6699ff}}

\title{Are Bias Mitigation Techniques for Deep Learning Effective?}

\author{Robik Shrestha$^1$ \quad Kushal Kafle$^{2}$ \quad Christopher Kanan$^{1,3,4}$ \\
Rochester Institute of Technology$^1$ \quad Adobe Research$^2$ \quad Paige$^3$ \quad Cornell Tech$^4$ \\
\texttt{\{rss9369, kk6055, kanan\}@rit.edu}
}


\maketitle

\begin{abstract}
   A critical problem in deep learning is that systems learn inappropriate biases, resulting in their inability to perform well on minority groups. This has led to the creation of multiple algorithms that endeavor to mitigate bias. However, it is not clear how effective these methods are. This is because study protocols differ among papers, systems are tested on datasets that fail to test many forms of bias, and systems have access to hidden knowledge or are tuned specifically to the test set. To address this, we introduce an improved evaluation protocol, sensible metrics, and a new dataset, which enables us to ask and answer critical questions about bias mitigation algorithms. We evaluate seven state-of-the-art algorithms using the same network architecture and hyperparameter selection policy across three benchmark datasets. We introduce a new dataset called \footnote{We recommend using new version of the dataset: BiasedMNISTv2 (https://github.com/erobic/occam-nets-v1)}{\gls*{BiasedMNIST}} that enables assessment of robustness to multiple bias sources. We use \gls*{BiasedMNIST} and a visual question answering (VQA) benchmark to assess robustness to hidden biases. Rather than only tuning to the test set distribution, we study robustness across different tuning distributions, which is critical because for many applications the test distribution may not be known during development. We find that algorithms exploit hidden biases, are unable to scale to multiple forms of bias, and are highly sensitive to the choice of tuning set. Based on our findings, we implore the community to adopt more rigorous assessment of future bias mitigation methods. \ifthenelse{\boolean{github}}{All data, code, and results are publicly available\footnote{https://github.com/erobic/bias-mitigators}.}{All data, code and results will be made publicly available.}
\end{abstract}

\section{Introduction}
\label{sec:intro}

Deep learning systems are trained to minimize their loss on a training dataset. However, datasets often contain spurious correlations and hidden biases which result in systems that have low loss on the training data distribution, but then fail to work appropriately on minority groups because they exploit and even amplify these spurious correlations~\cite{zhao2017men,10.3389/frai.2019.00028}.

\begin{figure}[ht]
    \centering
    \includegraphics[width=0.9\linewidth]{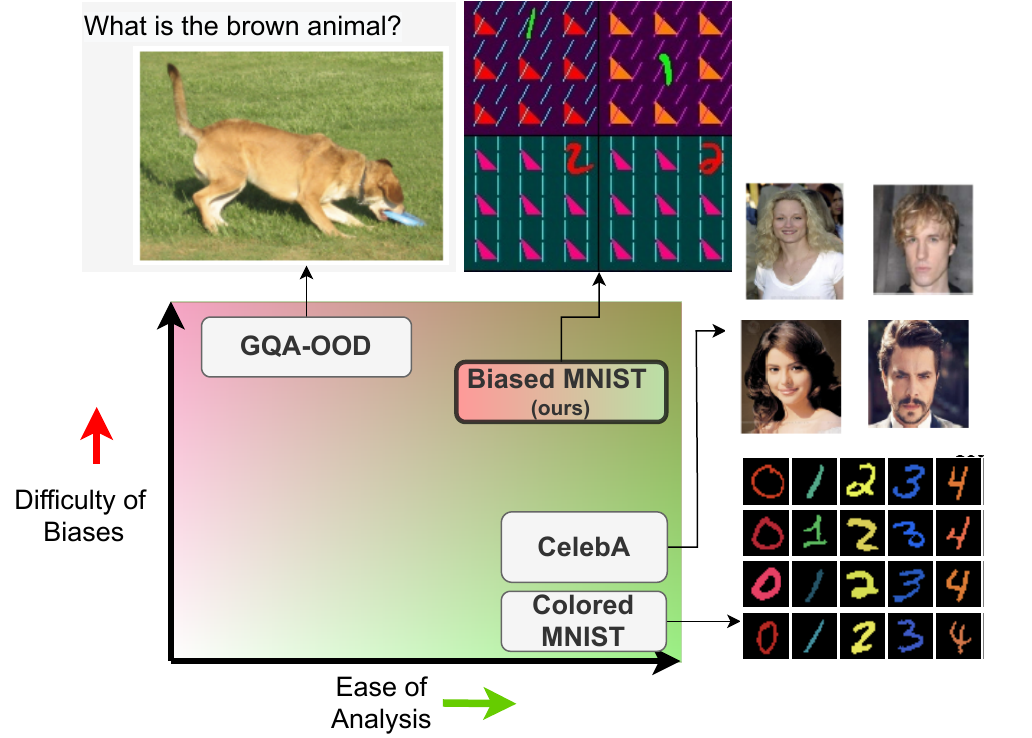}
    \includegraphics[width=0.9\linewidth]{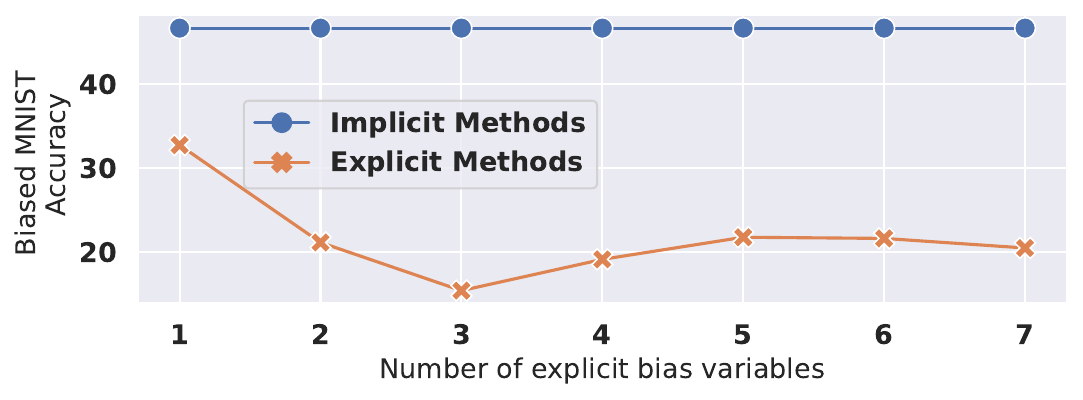}
    
    \caption{Current bias mitigation systems are tested on simple datasets that are easy to analyze, but do not offer challenges present in realistic cases. Addressing this, we propose the \gls*{BiasedMNIST} dataset which is easy to analyze, yet is reflective of real world challenges since it contains  multiple sources of biases. We find that methods fail on \gls*{BiasedMNIST} even when all the biases are explicitly labeled. We also test on GQA-OOD, where the sources of biases are not very obvious and, thus the methods fail to generalize.}
    
    \label{fig:datasets}
\end{figure}

\noindent For example, in systems trained to infer hair color on the CelebA dataset~\cite{liu2018large}, the majority group of non-blond males occurs $50$ times more than the minority group of blond males, resulting in systems incorrectly predicting non-blond as hair color for the minority group.

While this is a toy problem, in the real world, hidden minority patterns are common and failing on them can have dire consequences. Systems designed to aid human resources, help with medical diagnosis, determine probation, or loan qualification could be biased against minority groups based on age, gender, religion, sexual orientation, ethnicity, or race~\cite{SACHAN2020113100,barocas2016big,chouldechova2017fair,challen2019artificial,noseworthy2020assessing}. Systems can exploit correlated variables even if they are not directly a part of the input e.g., through inferred zip codes~\cite{datta2017use}, failing to work effectively on minority groups.

    

Recently, many methods have been proposed to make neural networks bias resistant. These methods can be grouped into two types: 1) those that assume the bias variables e.g., the gender label in CelebA, are explicitly annotated and can be accessed during training ~\cite{sagawa2019distributionally,sagawa2019distributionally,zhang2018mitigating,kim2019learning} and, 2) those that do not require  explicit access~\cite{nam2020learning,pezeshki2020gradient}. Assuming explicit access requires extra annotations in addition to the actual target, and for many tasks it may not be immediately clear what the bias variables are e.g., biases may only be discovered years later~\cite{pfungst1911clever,pezeshki2020gradient}. Methods that do not assume access to these bias variables have only recently been proposed~\cite{nam2020learning,utama2020towards,pezeshki2020gradient}.

So far, there is no study comparing methods from either group comprehensively. Often papers fail to compare against recent methods and vary widely in the protocols, datasets, architectures, and optimizers used. For instance, the widely used \gls*{ColoredMNIST} dataset, where colors and digits are spuriously correlated with each other, is setup differently across papers. Some use it as a binary classification task (class 0: digits 0-4, class 1: digits: 5-9)~\cite{arjovsky2019invariant,pezeshki2020gradient}, whereas others use a multi-class setting (10 classes)~\cite{kim2019learning,li2019repair}. For CelebA, \cite{nam2020learning} uses ResNet-18 whereas \cite{pezeshki2020gradient} uses ResNet-50, but the comparison was done without taking this architectural change into account. These discrepancies make it difficult to judge the methods on an even ground.
 
Methods are typically highly sensitive to hyperparameter choices, and papers report numbers on systems in which the hyperparameters were tuned using the test set distribution~\cite{clark2020learning,pezeshki2020gradient,teney2020value}. In the real world, biases may stem from multiple factors and may change in different environments, making this setup unrealistic. Furthermore, tuning on the test distribution can lead to methods that are right for the wrong reasons. When this is done, systems can perform well just by exploiting the biases they are supposed to overcome~\cite{shrestha-etal-2020-negative,teney2020value}, and they will then fail once deployed because they have not really learned to solve the task.

In addition, we posit that the commonly used benchmarks are not challenging enough to test generalization to realistic scenarios. For example CelebA and \gls*{ColoredMNIST}, two of the most widely used benchmarks, contain a single bias variable to mitigate: gender and color respectively. It is unclear how well methods would fare in presence of multiple types of bias, e.g., position or co-occurring objects/patterns, which are commonly present in real-world datasets. For some tasks it can be impossible to exhaustively enumerate all bias variables. For example, in visual question answering (VQA), where a system answers questions about images, biases can stem from: object-context co-occurrences, visual concept/language correlations, question type/answer distributions, and more. Annotating all such sources of bias is unrealistic. Even when the bias variables are explicitly labeled, it is still unclear if the methods can remain robust to all of the bias sources, since this entails generalization to a large number of dataset groups e.g., hundred thousand groups for GQA-OOD~\cite{kervadec2020roses}.


\paragraph{We address the above issues via these contributions:}
\begin{enumerate}[noitemsep, nolistsep]
    \item We describe our new \gls*{BiasedMNIST}  dataset and corresponding evaluation protocol for measuring resistance to multiple forms of bias. It measures resistance to spuriously correlated background/foreground color, texture, co-occurring distractors, position, and more.
    \item We compare seven state-of-the-art bias mitigation methods on classification tasks using \gls*{BiasedMNIST} and CelebA, measuring generalization to minority patterns, scalability to multiple sources of biases, sensitivity to hyperparameters, etc. We ensure fair comparisons by using the same architecture, optimizer, and performing grid searches for hyperparameters.
    \item To go beyond image classification, we measure the performance of these methods on the biased GQA benchmark for VQA. 
    \item We provide concrete recommendations for future studies, so that comparisons among algorithms are meaningful and reflective of real-world challenges.
\end{enumerate}

\section{Problem Statement}
To properly study bias mitigation, it is necessary to provide a definition of biased data and biased behavior in a model. We study bias in supervised classification i.e., the goal is to learn a function $f : X \rightarrow Y$ which outputs a categorical target $y \in Y$ given $x \in X$. Each $x$ is itself a mixture of a signal $s$ that we wish the system to use for inference and bias $b$ that is spuriously correlated with $y$. Since the spurious correlations between $y$ and $b$ do not always hold, systems exploiting $b$ to infer $y$ fail to generalize. 

We can measure the robustness to such tendencies by intentionally introducing covariate shift e.g., with a test dataset distribution that differs from training or a metric that balances performance across groups. For our study, we use the mean per group accuracy/unbiased accuracy, which weighs all the groups equally. Furthermore, we focus on the cross-bias setting defined by~\cite{bahng2020learning} where the same set of bias variables are present in both train and test sets.



\section{Bias Mitigation Strategies}
\label{sec:strategies}

Without bias mitigation mechanisms, \textbf{standard models (\gls*{conventional})} often use spurious bias variables for inference, rather than developing invariance to them, which often results in their inability to perform well on minority patterns~\cite{he2009learning,bolukbasi2016man,agrawal2018don,shrestha2019answer}. To address this, several bias mitigation mechanisms have been proposed, and they can be categorized into two groups: 1) methods that access explicit bias labels during training, and 2) methods that do not assume such access. We briefly review methods from these categories, with an emphasis on the methods assessed in our studies.

\subsection{Explicit Bias Mitigation Methods}
Explicit bias mitigation techniques directly access the bias variables: $b_{expl.}$ during training to develop invariance to them. Based on the way these variables are utilized during training, we choose five different explicit methods for our study. We refer to them as \textbf{explicit methods} for conciseness.

\textbf{Re-sampling/Re-weighting:} These approaches balance out the spurious correlations. The classical approach is to re-balance the class distribution by adjusting the sampling probability/ loss weight for majority/minority samples~\cite{chawla2002smote,he2008adasyn,lin2017focal,zou2018unsupervised,cui2019class}. This includes synthesizing minority instances too~\cite{chawla2002smote,he2008adasyn}. Moving beyond class imbalances, REPAIR~\cite{li2019repair} proposed learning dynamic weights to mitigate representation bias~\cite{li2018resound}. However,~\cite{sagawa2019distributionally} have shown promising results by using static weights to upweight minority patterns. We choose this method due to its simplicity.

\textit{Group Upweighting (\gls*{Up Wt})~\cite{sagawa2019distributionally}} attempts to mitigate the correlations between $y$ and $b_{expl.}$ by upweighting the minority patterns. Specifically, each sample ($x, y$) is assigned to a group: $g=(y, b_1, b_2, .., b_E)$, where $E$ is the total number of variables contained in $b_{expl.}$ and the loss is scaled by $\frac{1}{N_g}$, where $N_g$ is the number of instances in group $g$. Up Wt requires the models to be sufficiently regularized, i.e., be trained with low learning rates and/or high weight decays to be robust to the minority groups.

\textbf{Distributionally Robust Optimization (DRO):} DRO~\cite{delage2010distributionally} minimizes the worst-case expected loss over potential test distributions. Often, such distributions are approximated by sampling from a uniform divergence ball around the train distribution~\cite{ben2013robust,duchi2021statistics,namkoong2016stochastic}. However, this lacks structured priors about the potential shifts, and instead hurts generalization~\cite{hu2018does}.

\textit{Group DRO (\gls*{GDRO})~\cite{sagawa2019distributionally}} provides DRO with the necessary prior that it must generalize to all groups. Similar to Up Wt, GDRO also uses $y$ and $b_{expl.}$ to create groups and has been shown to work well with sufficiently regularized models. However, unlike Up Wt, it performs weighted sampling from each group and has an optimization procedure to minimize the loss over the worst-case group.

\textbf{Ensembling Approaches:}
Ensembling approaches~\cite{he2019unlearn,clark2019don,cadene2019rubi} have a two-branch setup: a) a bias-only branch $f_{b}$ that predicts $y$ from $b$ alone to identify the bias-prone samples, and b) a de-biased branch $f_d(.)$ that is trained to focus on samples that $f_b$ finds difficult so that it learns richer features that work on difficult samples too. The two branches can be ensembled in different ways. DRiFt~\cite{he2019unlearn} uses product-of-experts~\cite{hinton2002training} and LearnedMixin~\cite{clark2019don} extends this through learned weights and entropy constraints that control $f_b$. 

\textit{Reduction of Unimodal Biases (\gls*{RUBi}) ~\cite{cadene2019rubi}} multiplies the outputs from $f_{d}(.)$ with sigmoided outputs from $f_{b}(.)$, thereby assigning higher loss weights to samples that cannot be predicted through biases alone. RUBi was the previous state-of-the-art on VQA-CP~\cite{agrawal2018don}, a testbed for measuring robustness to biases in VQA. For the bimodal problem of VQA, the original implementation focused on linguistic biases, training $f_b$ on question features only. For our studies, we instead train $f_b$ on $b_{expl.}$ directly, to control the type of biases captured by $f_b$. We assess RUBi~\cite{cadene2019rubi} over others since it performed better in the preliminary studies.

\textbf{Adversarial Debiasing:}
These techniques impair the ability of the representation learner to encode biases~\cite{zhang2018mitigating,adeli2019bias,ramakrishnan2018overcoming,grand2019adversarial}. Like ensembling methods, they also employ a two-branch setup, with the representation encoder in the main branch being penalized if the bias-only branch: $f_b()$ is successful at predicting biases from them~\cite{zhang2018mitigating}. Alternately, $f_b()$ may be trained to predict the class label from the biased features ~\cite{ramakrishnan2018overcoming,grand2019adversarial}, but in either case, the gradient from $f_b()$ is reversed during backpropagation for debiasing.

\textit{Learning Not to Learn (\gls*{LNL}) \cite{kim2019learning}} uses an adversarial setup derived from minimization of mutual information between representation and bias. In addition to the gradient reversal, the mutual information formulation introduces an entropy regularization on the bias predictions.

\textbf{Invariant Risk Minimization (IRM):} The goal of IRM is to extract representations that are invariant across environments: $\mathcal{E} = \{e_1, e_2,...e_E\}$, each encoding different spurious correlations~\cite{arjovsky2019invariant,teney2020unshuffling,choe2020empirical}. Such representations enable the same classifier to be simultaneously optimal over all $\mathcal{E}$. For this, ~\cite{arjovsky2019invariant} propose to regularize the gradient norm of a fixed linear classifier. More recent variants include regularization of variance of risks~\cite{krueger2020out,xie2020risk}. However, ~\cite{rosenfeld2020risks} have shown that such objectives can fail to recover the invariant features in practice. Despite this negative result, we still compare against IRM since it is a promising research direction.
 
\textit{\gls*{IRMv1}~\cite{arjovsky2019invariant}}  is an efficient approximation of an otherwise computationally expensive bi-level IRM objective. It consists of a regularization constraint on the gradient norm with respect to a fixed scalar $\theta_c = 1.0$:
\begin{align*}
 \underset{\theta}{min} \sum_{e \in \mathcal{E}_{tr}} l^e(\hat{y})    + \lambda ||\nabla_{\theta_c|\theta_c=1.0} l^e(\theta_c.\hat{y})||^2,
\end{align*}
where, $l^e$ is loss on environment $e$, $\hat{y}$ is the logit vector yielded by the model parameterized by $\theta$ and $\lambda$ balances between the empirical risk and invariance. In our experiments, we use the previously defined explicit data groups as the training environments for IRM.

\subsection{Implicit Bias Mitigation Methods}

Since explicit access to bias variables is an undesirable requirement in practice, some recent methods have proposed to mitigate biases without such assumptions. We call them \textbf{implicit methods} for conciseness and describe them below.

\textbf{Limited Capacity Models:}
Most implicit methods assume that easy-to-learn biases can be captured by limiting the capacity of the models~\cite{utama2020towards,sanh2020learning,nam2020learning}. The capacity of such bias-prone models: $f_b()$ can be limited by using a small subset of train instances for a few epochs~\cite{utama2020towards}, using fewer model parameters~\cite{sanh2020learning}, attaching a classifier to intermediate layers, instead of using the final representation layers~\cite{clark2020learning}, using bias-prone architectures~\cite{bahng2019learning} or amplifying biases~\cite{nam2020learning}. Main network: $f_d$ is then debiased by assigning higher weights to the harder samples, so that it generalizes to samples that cannot be predicted through biases alone.

\textit{Learning From Failure (\gls*{LFF})~\cite{nam2020learning}} amplifies the bias in $f_b$ using the generalized cross entropy loss~\cite{zhang2018generalized}:
\begin{align*}
    GCE(s(x;\theta), y) = \frac{1 - s_y(x; \theta)^\gamma}{\gamma},
\end{align*}
where $s_y$ is the softmax score for the ground truth class and $\gamma$ determines the degree of bias amplification. Samples with high $f_b$ loss are then assigned higher weights while training $f_d$. While $\gamma$ seems critical, the original paper does not discuss a way to tune it and instead fixes it to a default value of $\gamma=0.7$. In fact, the paper does not provide details on model selection at all; however, this is an important question, which we discuss in Sec.~\ref{sec:model_selection}.

\textbf{Gradient Starvation Mitigation:}
Different from the limited capacity methods, spectral decoupling~\cite{pezeshki2020gradient} aims to overcome the issue of gradient starvation~\cite{combes2018learning}, which is the tendency to only rely on statistically dominant features. This is related to the simplicity bias exploitation, where models exploit the simplest features despite having access to more predictive features~\cite{shah2020pitfalls,hermann2020shapes}, which are more complex.

\textit{Spectral Decoupling (\gls*{SD})~\cite{pezeshki2020gradient}} aims to decouple the learning dynamics between features. The authors show that regularizing the network outputs ($\hat{y}$) as:
\begin{align*}
    \frac{\lambda}{2}||\hat{y} - \gamma||_2^2,
\end{align*}
where, $\lambda$ and $\gamma$ are hyperparameters, provably decouples the learning dynamics, enabling learning of better features.

\section{Datasets}
\label{sec:datasets}
We use datasets that enable probing existing methods with critical questions regarding their robustness. We test on datasets with varying scales and types of biases, allowing us to perform highly controlled studies that analyze scalability to a large number of hidden groups.

\subsection{Biased MNIST}

\begin{figure}[ht]
    \centering
    \includegraphics[width=0.3\textwidth]{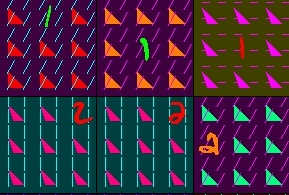}
    \caption{\gls*{BiasedMNIST} requires the methods to classify the target digit while remaining invariant to multiple sources of biases.}
    \label{fig:biased_mnist_composite}
\end{figure}

Existing datasets for assessing bias mitigation methods do not enable analysis of multiple bias sources, e.g., \gls*{ColoredMNIST} only tests for color versus class bias. To address this, we created the Biased MNIST dataset, which requires recognizing digits while remaining robust to multiple sources of biases i.e., other factors which are also correlated with the target variable. Specifically, the dataset consists of $3 \times 3$ grids of cells, where the target digit is placed at one of the grid cells and is correlated with multiple bias variables $\{b_j\}$ including: a) background color, b) digit color, c) digit position in the grid, d) distracting shapes present in other cells, e) color of the distracting shapes, f) texture type and g) texture color (see Fig.~\ref{fig:biased_mnist_composite}). Each variable can take one of ten discrete values: each variable takes the majority/biased value corresponding to the target digit with a probability of $p_{bias}$, otherwise takes one of the remaining nine values with uniform probability. For instance, if $p_{bias} = 0.7$, then $70\%$ of 1's are colored as green, $70\%$ of 2's are colored red and so on. This applies to all the variables e.g., $70\%$ of 1's placed on a purple background, while $70\%$ of 2's are placed on a green background. The bias level: $p_{bias}$ can be specified for each variable to control the types and degrees of biases. For convenience, $p_{bias}$ is set to $0.7$ in the train set, unless otherwise specified. The test set is unbiased i.e., has $p_{bias}=0.1$.


\subsection{CelebA}
The CelebA dataset~\cite{liu2018large} of celebrity faces is widely used to assess bias mitigation techniques~\cite{sagawa2019distributionally,sagawa2020investigation,nam2020learning,pezeshki2020gradient}. Following earlier work, it is used for binary hair color classification (blond or non-blond), which is correlated with gender. There are two major bias sources: a) class imbalance, with non-blond  occurring 5.7 times more than blond hair color, and b) presence of a rare group, i.e., blond male celebrities only account for 0.86\% of the training instances. 

\subsection{GQA-OOD}
\label{sec:gqa_ood}
We use the GQA visual question answering dataset~\cite{hudson2019gqa} to highlight the challenges of using bias mitigation methods on real-world tasks. It has multiple sources of biases including imbalances in answer distribution, visual concept co-occurrences, question word correlations, and question type/answer distribution. It is unclear how the explicit bias variables should be defined so that the methods can generalize to all minority groups. GQA-OOD~\cite{kervadec2020roses} divides the evaluation and test sets into majority (head) and minority (tail) groups based on the answer frequency within each `local group' (e.g., colors of bags), which is a unique combination of `global group' or answer type (e.g., objects or colors) and the main concept asked in the question (e.g., `bag', `chair', etc.). The head/tail categorization makes analysis easier; however, it is unclear how one should specify the explicit biases so that the models generalize even to the rarest of local groups. Therefore, we explore multiple ways of defining the explicit bias variable in separate experiments: a) majority/minority group label (2 groups), b) answer class (1833 groups), c) global group (115 groups) and d) local group (133328 groups). It is unknown if bias mitigation methods can scale to hundreds and thousands of groups in GQA, yet natural tasks require such an ability.

\section{Network Architecture \& Tuning Procedure}

For each dataset, we assess all bias mitigation methods with the same neural network architecture. For CelebA, we use ResNet-18~\cite{he2016deep}. For \gls*{BiasedMNIST}, we use a convolutional neural network with four ReLU layers consisting of a max pooling layer attached after the first convolutional layer. For GQA-OOD, we employ the UpDn architecture~\cite{anderson2018bottom}, which is widely used for VQA~\cite{selvaraju2019taking,kervadec2020roses, wu2020improving}.

For each dataset, we use the class label $y$ and the explicit bias variables $b_{expl.}$ to define explicit groups for Up Wt, GDRO and IRMv1. For instance, for CelebA, hair color and gender result in four explicit groups while for \gls*{BiasedMNIST}, the number of groups: $|G|$ depends on the number of explicit bias variables: $|b_{expl.}|$, with $|G| = 10^{|b_{expl.}|}$. We will specify the exact $b_{expl.}$ for each experiment in Sec.~\ref{sec:analysis}. For GQA, we use head/tail, answer class, global and local groups as explicit variables. For all datasets, RUBi uses $b_{expl.}$ to predict $y$, whereas LNL trains the adversarial branch to predict $b_{expl.}$ from representations. Of course the implicit methods: \gls*{conventional}, LFF and SD are invariant to the choice of the explicit biases during training.  Unless otherwise specified, results from \gls*{BiasedMNIST} are averaged across 3 random seeds, but due to computational constraints, we ran models on CelebA and GQA-OOD only once.

Hyperparameters for each method were chosen using a grid search with unbiased accuracy on each dataset's validation set. To make this tractable, we first ran a grid search for the learning rate over $\{10^{-3}, 10^{-4}, 10^{-5}\}$ and weight decay over $\{0.1, 10^{-3}, 10^{-5},0\}$. After the best values were chosen, we searched for method-specific hyperparameters. Due to the size of GQA-OOD, hyperparameter search was performed by training on only 10\% of instances, and then the best selected hyperparameters were used with the full training dataset. The exact values for the hyperparameters are specified in the Appendix. 

\section{Questions Posed and Answered}

\label{sec:analysis}

In this section, we probe the existing methods with critical questions regarding their robustness. For each question, we first describe the empirical setup to explore the question, and then present the results.

\subsection{Head-to-Head Comparisons}

\label{sec:head_to_head}

\textit{Question 1: Are there clear winners in a head-to-head comparisons across datasets?}
\label{qn1}

We first present the mean per group accuracy for all eight methods on all three datasets in Table.~\ref{tbl:overall-results} to see if any method does consistently well across benchmarks. For this, we used class and gender labels as explicit biases for CelebA. For \gls*{BiasedMNIST}, there are multiple ways to define explicit biases, but for this section, we simply use each of the seven variables as explicit biases in different runs and average across the runs. We study combinations of multiple explicit variables in Sec.~\ref{sec:scalability_of_methods}. We set $p_{bias}=0.7$ for this section, and present results across different $p_{bias}$ in the Appendix. Similarly for GQA, we consider each of the four variables described in Sec. ~\ref{sec:gqa_ood} as explicit bias in separate runs and present the average.

\begin{table}[]
\centering
\caption{Unbiased accuracies $Acc(\alpha=0)$ on all datasets for all methods. We format the $\textbf{\underline{\textit{first}}}$, $\textbf{second}$ and $\underline{third}$ best results. Methods that do not access explicit biases have gray background.}
\label{tbl:overall-results}
\begin{tabular}{lccc}
\hline
\begin{tabular}[c]{@{}l@{}}Methods/\\ Datasets\end{tabular} & CelebA &  \begin{tabular}[c]{@{}l@{}}Biased MNIST \end{tabular} & GQA \\ \hline
\rowcolor[HTML]{D0D0D0} 
\gls*{conventional}  & 80.3  & \textbf{42.0} & \underline{44.8}   \\
Up Wt~\cite{sagawa2020investigation} & \underline{87.4} & 30.1  & 30.0 \\
GDRO~\cite{sagawa2019distributionally}  & \textbf{88.5} & 27.2  & 26.4 \\
RUBi~\cite{cadene2019rubi}  &  87.2 & 38.9 & 24.1 \\
LNL~\cite{kim2019learning}   &  79.2 & 40.6 & 28.6 \\
IRMv1~\cite{arjovsky2019invariant} & 79.8  & 38.7 & 39.3 \\
\rowcolor[HTML]{D0D0D0} 
LFF~\cite{nam2020learning}   & 77.8 & \textbf{\underline{\textit{56.6}}} & \textbf{45.1} \\
\rowcolor[HTML]{D0D0D0} 
SD~\cite{pezeshki2020gradient} & \textbf{\underline{\textit{88.6}}} & \underline{41.3} & \textbf{\underline{\textit{46.9}}} \\ \hline
\end{tabular}%
\end{table}

\textbf{Results.} As shown in Table.~\ref{tbl:overall-results}, no method performs universally well across datasets; however, the implicit methods LFF and SD obtain high unbiased accuracies on most datasets. This shows that implicit methods can deal with multiple bias sources without explicit access. Explicit methods work well on CelebA but fail on \gls*{BiasedMNIST} and GQA. Specifically, Up Wt, GDRO and RUBi obtain 7-8\% improvements over \gls*{conventional} on CelebA, which requires generalization to only 4 groups. However, all explicit methods perform worse than \gls*{conventional} on \gls*{BiasedMNIST} and GQA, signifying their inability to deal with multiple bias sources. LNL and IRMv1  were comparable to \gls*{conventional} even on CelebA, demonstrating lack of generalization even on simple settings. Despite being a simpler method, Up Wt outperformed GDRO on both \gls*{BiasedMNIST} and GQA, but both were worse than \gls*{conventional}. These results show that implicit methods can outperform explicit methods.

\subsection{Bias Exploitation}
\label{sec:bias_exploitation}

\textit{Question 2: Do methods show robustness to both explicit and implicit biases?}

In this set of experiments, we compare the resistance to explicit and implicit biases. We primarily focus on the \gls*{BiasedMNIST} dataset, reserving each individual variable as the explicit bias in separate runs of the explicit methods, while treating the remaining variables as implicit biases. To ease analysis, we compute the accuracy gap between the majority and minority groups i.e., majority/minority difference (MMD). Majority/minority groups are defined per variable e.g., for foreground color, green 1's, red 2's etc are placed in the majority group and the rest in the minority group and MMD simply computes the accuracy difference between the two groups for each variable. High MMDs indicate that the methods rely heavily on spurious patterns favoring the majority groups and thus fail on the minority groups.

\begin{figure}[t]
    \centering
    \subcaptionbox{\label{fig:biased_mnist_variable_mmd_boxplot}}
    {\includegraphics[width=0.45\textwidth]{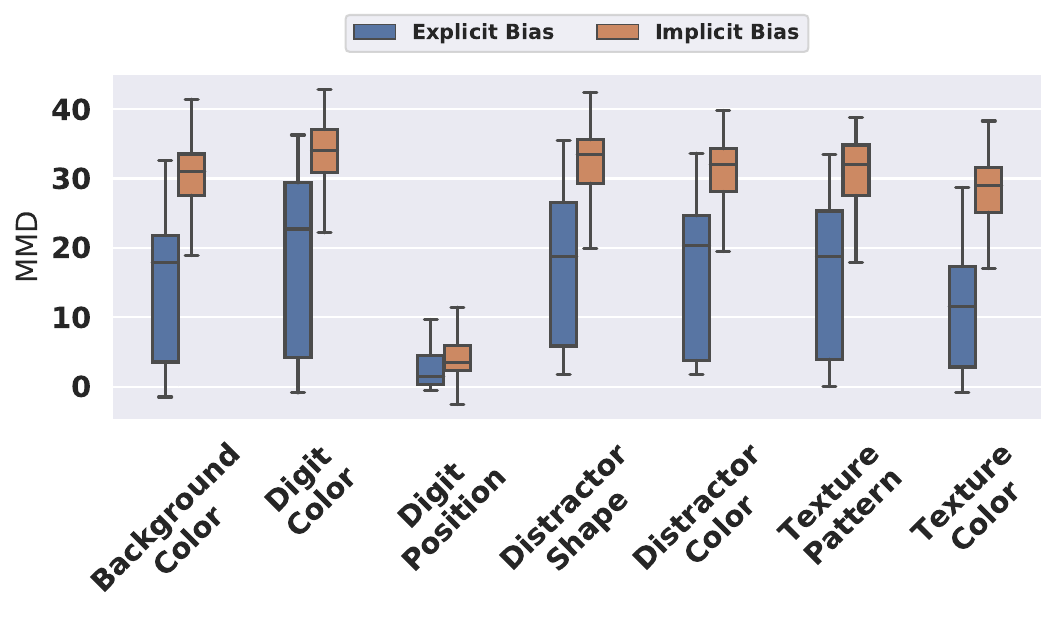}}
    \subcaptionbox{\label{fig:biased_mnist_methods_mmd_boxplot} }
    {\includegraphics[width=0.45\textwidth]{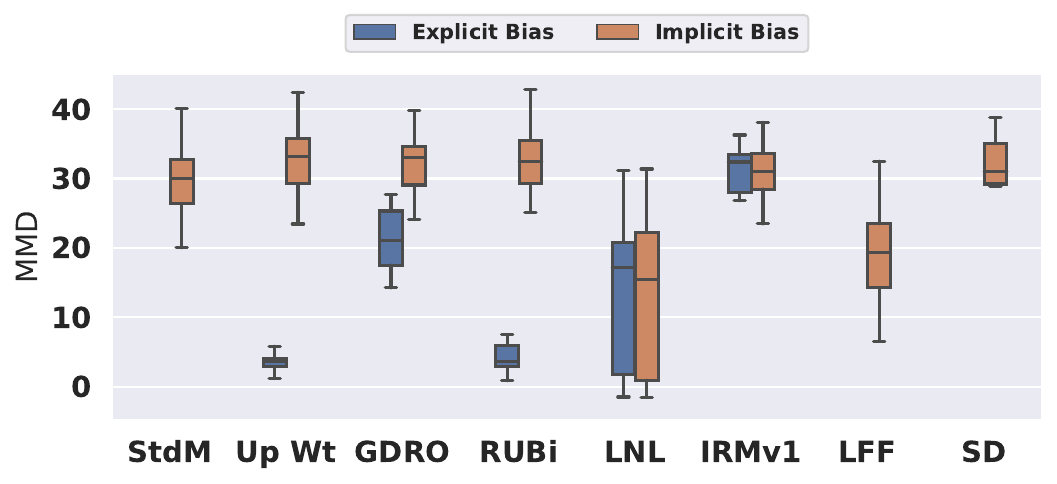}}
    \caption{Boxplots of majority/minority difference (MMD) on \gls*{BiasedMNIST} over: a) bias variables, and b) different methods.}
    \label{fig:biased_mnist_mmd}
\end{figure}

\textbf{Results.}
In Fig.~\ref{fig:biased_mnist_variable_mmd_boxplot}, we present the MMD boxplots for all bias variables, comparing cases when the label of the variable is either explicitly specified (explicit bias), or kept hidden (implicit bias) from the methods. Barring digit position, we observe that the MMD values are higher when the variables are not explicitly labeled for the methods, indicating that the explicit methods in general fail to mitigate implicit biases. Fig.~\ref{fig:biased_mnist_methods_mmd_boxplot} breaks down exploitation of explicit and implicit biases for each method. UpWt, GDRO and RUBi have low MMD values for explicit biases, but high MMD values for implicit biases, showing that they mitigate the explicit biases to some extent, but are not robust to the implicit biases. LNL and IRMv1 seem to be equanimously affected by both explicit and implicit biases, and thus fail to improve upon the baseline as previously shown in Table~\ref{tbl:overall-results}. LFF has a relatively low range of MMDs and as shown by the improvements in Table~\ref{tbl:overall-results}, the method outperforms others on Biased MNIST.



\ifthenelse{\boolean{coordconv}}{
Interestingly, MMD was low for digit position. We hypothesize this is because CNNs are unable to use position information for inference~\cite{liu2018intriguing}. To confirm this, we add CoordConv layers~\cite{liu2018intriguing} before and after the maxpooling layer in CNN to enable usage of position information. This resulted in methods exploiting digit position too, showing larger MMD values of 11.1\%-25.6\% as compared to the 2.2\%-8.7\% without the CoordConv layers. Such inductive biases affect whether or not methods exploit certain dataset biases, and we discuss this in Sec.~\ref{sec:discussions}.
}


\subsection{Scalability of Methods}
\label{sec:scalability_of_methods}
\begin{figure}[t]
    \centering
    \includegraphics[width=0.4\textwidth]{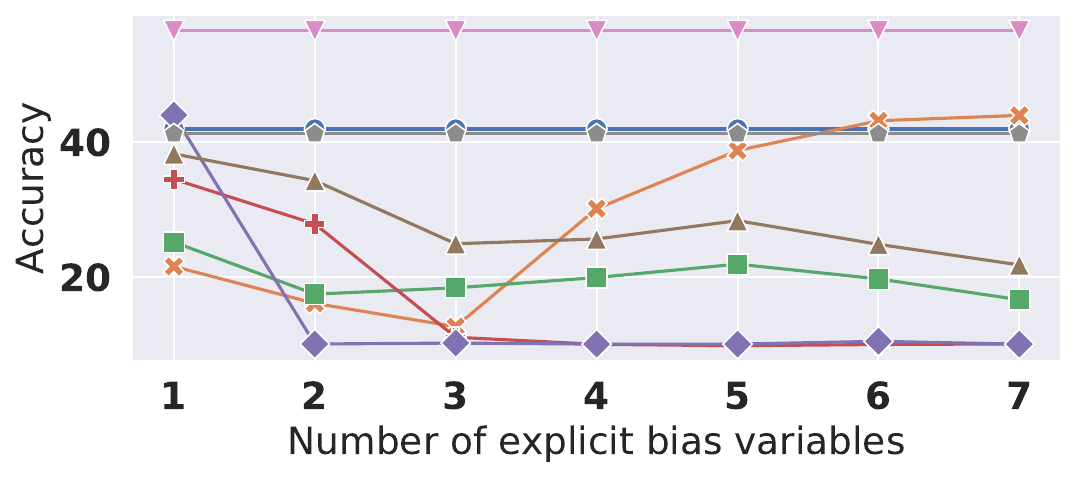}
    \includegraphics[width=0.35\textwidth]{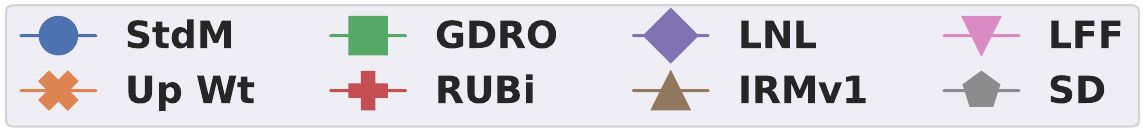}
    \caption{Unbiased accuracy as a function of number of explicit bias variables. \gls*{conventional}, LFF and SD are invariant to the choice of explicit bias variables.}
    \label{fig:biased_mnist_increasing_variables}
\end{figure}

\begin{table}[]
\centering
\caption{Mean of head and tail accuracies on GQA, considering different variables as explicit biases.}
\label{tbl:gqa_expl_bias}
\begin{tabular}{lcccc} \hline
Methods &
  \begin{tabular}[c]{@{}c@{}}Head/Tail\\ (2 \\ groups)\end{tabular} &
  \begin{tabular}[c]{@{}c@{}}Answer\\ Class\\ (1833\\ groups)\end{tabular} &
  \begin{tabular}[c]{@{}c@{}}Global\\ Group\\ (115 \\ groups)\end{tabular} &
  \begin{tabular}[c]{@{}c@{}}Local\\ Group\\ (133328\\ groups)\end{tabular} \\ \hline
\rowcolor[HTML]{D0D0D0} 
\gls*{conventional}  & \multicolumn{4}{c}{44.8}  \\
Up Wt~\cite{sagawa2020investigation} & 43.3 & 26.0 & 26.4 & 24.2 \\
GDRO~\cite{sagawa2019distributionally}  & 46.9 & 28.6 & 10.8 & 19.4 \\
RUBi~\cite{cadene2019rubi}  & 44.1 & N/A  & 5.6  & 22.6 \\
LNL~\cite{kim2019learning}   & 42.9 & N/A  & 32.4 & 10.7 \\
IRMv1~\cite{arjovsky2019invariant} & 47.2 & 35.8 & 40.4 & 33.8 \\
\rowcolor[HTML]{D0D0D0} 
LFF~\cite{nam2020learning}   & \multicolumn{4}{c}{45.1}  \\
\rowcolor[HTML]{D0D0D0} 
SD~\cite{pezeshki2020gradient}    & \multicolumn{4}{c}{46.9} \\ \hline
\end{tabular}%
\end{table}
\textit{Question 3: Do methods scale up to multiple types of biases and a large number of dataset groups?}

It is unknown how well the methods scale up to multiple sources of biases and large number of groups, even when they are explicitly annotated. To study this, we train the explicit methods with multiple explicit variables for \gls*{BiasedMNIST} and individual variables that lead to hundreds and thousands of groups for GQA and compare them with the implicit methods. For \gls*{BiasedMNIST}, we first sort the seven total variables in the descending order of MMD (obtained by \gls*{conventional}) and then conduct a series of experiments. In the first experiment, the most exploited variable, distractor shape, is used as the explicit bias. In the second experiment, the two most exploited variables, distractor shape and texture, are used as explicit biases. This is repeated until all seven variables are used\footnote{The exact order is given in the Appendix.}. Note that conducting the seventh experiment entails annotating each instance with every possible source of bias. While this may not be realistic in practice, such a controlled setup will reveal if the explicit methods can generalize when they have complete information about every bias source. 


To test scalability on a natural dataset, we conduct four experiments per explicit method on GQA-OOD with the explicit bias variables: a) head/tail (2 groups), b) answer class (1833 groups), c) global group (115 groups), and d) local group (133328 groups). Unlike \gls*{BiasedMNIST}, we do not test with combinations of these variables since the last three variables already entail generalization to many groups.

\textbf{Results.} We find that implicit methods either improve or are comparable with \gls*{conventional}, but most explicit methods fail when asked to generalize to multiple bias variables and a large number of groups, even when the bias variables are explicitly provided. As shown in Fig.~\ref{fig:biased_mnist_increasing_variables}, all explicit methods are below \gls*{conventional} on \gls*{BiasedMNIST}. Barring LNL and Up Wt, other explicit methods exhibit degraded accuracy as the number of explicit bias variables increases. Because the implicit methods do not rely on the choice of explicit biases, we simply repeat the same accuracy across x-axis. Among the implicit methods, LFF obtains the highest improvement, whereas SD is close to \gls*{conventional}.

Results for GQA-OOD are similar, with explicit methods failing to scale up to a large number of groups, while implicit methods showing some improvements over \gls*{conventional}. As shown in Table~\ref{tbl:gqa_expl_bias}, when the number of groups is small, i.e., when using a head/tail binary indicator as the explicit bias, explicit methods remain comparable or even outperform \gls*{conventional}, but when the number of groups grow to hundreds and thousands, they fail. IRMv1 and GDRO obtain the highest improvements of 2.4\% and 2.1\% over \gls*{conventional}, respectively, with the binary head/tail bias, but they show large drops when using answer class, global group or local group as explicit bias variables. Some drops are extreme, e.g., RUBi  drops 39\%  when using global group as the explicit bias variable. 

\begin{figure}[t]
    \centering
    \includegraphics[width=0.45\textwidth]{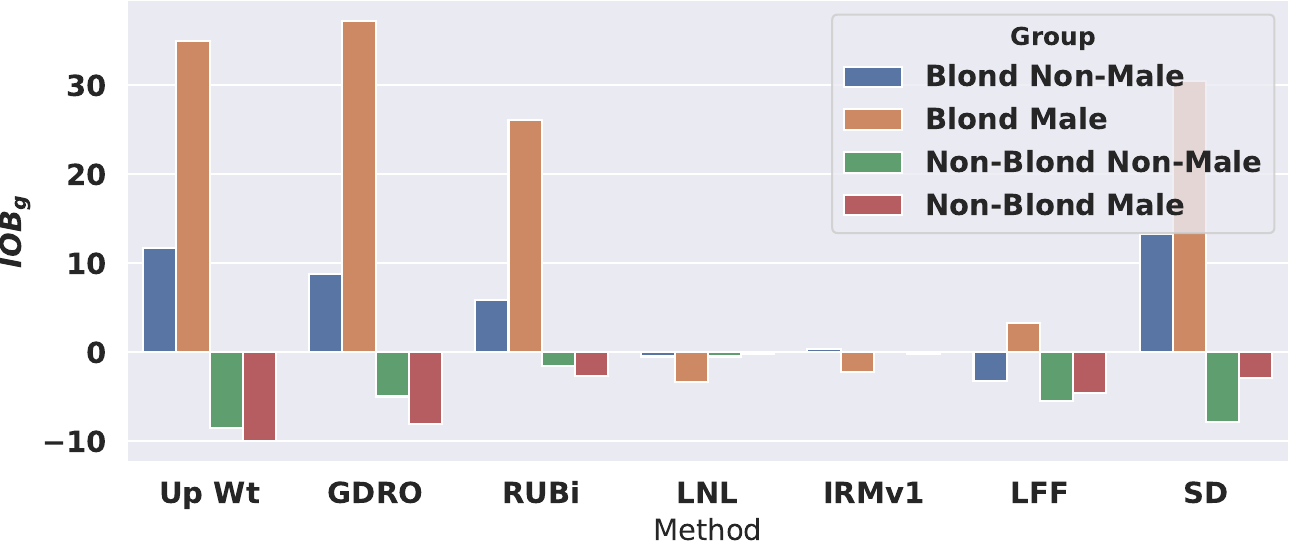}
    \caption{Improvement Over the Standard Model (IOSM) for each group of CelebA.}
    \label{fig:CelebA_delta_without_priors}
\end{figure}

\textbf{Results on a simpler setting.} We further study bias exploitation on CelebA. For this, we plot improvement over the standard model ($IOSM$) in Fig.~\ref{fig:CelebA_delta_without_priors}, which is the accuracy gain over the standard model on each dataset group. The improvements in  blond (minority group) incur degradation in non-blond (majority group). The methods tilt predictions either in the favor of minority or majority groups, which shows the inability to learn the signal even on simple settings.

\subsection{Robustness to Model Selection Criteria}
\label{sec:model_selection}
\textit{Question 4: Can the methods generalize to the test set without being tuned on the test distribution? Do they exhibit robustness across a wide range of hyperparameters?}

Assuming access to the test distribution for model selection is unrealistic and can result in models being right for the wrong reasons~\cite{teney2020value}. Rather, it is ideal if the methods can generalize without being tuned on the test distribution and we study this ability by comparing models selected through varying tuning distributions. To control the tuning distribution, we define a generalization of the mean per group accuracy (MPG) metric, that can interpolate within as well as extrapolate beyond the train and test distributions:
\begin{align*}
    Acc(\alpha) = \frac{\sum_{g=1}^{|G|} p_{g}^{\alpha} Acc_g}{\sum_{g=1}^{|G|} p_{g}^{\alpha}}.
\end{align*}
Here, $p_{g}$ denotes the ratio of samples present in group $g$, $|G|$ is the total number of groups and $\alpha$ is used to control the group prior. When $\alpha=0$, $Acc(\alpha=0)$ yields the MPG i.e., a balanced distribution where all groups are weighed equally. When $\alpha=1$, then the weights reflect the train priors/biases. When $0<\alpha<1$, it interpolates between biased (train) priors and unbiased group weights. When $\alpha<0$, minority groups are weighed more and when $\alpha > 1$, majority groups are weighed more i.e., it amplifies the train bias. 

Ideally, methods should yield robust models regardless of the tuning distribution i.e., the value of $\alpha$. To test this ability, we train a set of candidate models for model selection, with different sets of hyperparameters. Specifically, we train nine different models with $\textit{learning rate} \in \{10^{-3}, 10^{-4}, 10^{-5}\}$ and $\textit{weight decay} \in \{10^{-1}, 10^{-3}, 10^{-5}\}$ for CelebA and \gls*{BiasedMNIST} and, then perform model selection by computing $Acc(\alpha)$ at $\alpha \in \{-1.0, -0.5, 0.0, 0.5, 1.0, 1.5, 2.0\}$ on the validation sets.

\begin{figure}[t]
    \centering
    \includegraphics[width=0.4\textwidth]{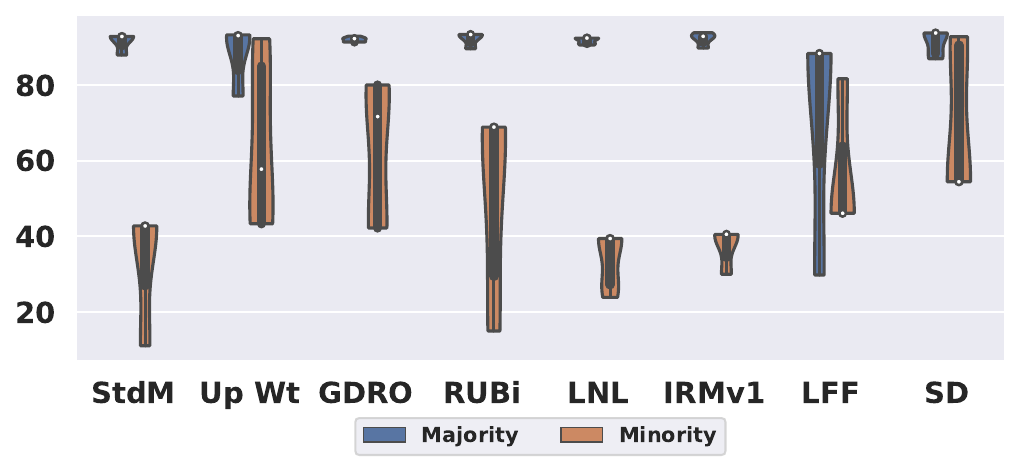}
    \caption{Range of minority (blond haired males) and majority (mean over rest of the groups) test accuracies on CelebA when varying the validation distribution from $\alpha=-1$ (inverted train bias) to $\alpha=2$ (increased train bias).} 
    \label{fig:CelebA_model_selection_maj_min}
\end{figure}

\textbf{Results.}
For CelebA, methods generally show large variance on the minority patterns (blond haired male celebrities), and lower variance on the majority patterns (mean over rest of the groups), whereas for \gls*{BiasedMNIST}, we find that methods only work for certain set of hyperparameters and show degraded results on both majority/minority patterns if the hyperparameters change. As illustrated by the violin plots of CelebA's unbiased test accuracies in Fig.~\ref{fig:CelebA_model_selection_maj_min},  LNL and IRMv1 have the lowest variance, but neither improves over \gls*{conventional}. Up Wt, GDRO and RUBi show the largest variances for the minority group, indicating they are highly sensitive to the choice of tuning distribution. For LFF, we found high variance for both majority and minority patterns. In fact, we were unable to replicate the published LFF results, with $\gamma=0.7$ yielding a high accuracy ($86\%$) on the rarest group, but low accuracies on the rest. After tuning it, we found that $\gamma=0.1$ gave the best unbiased test accuracy.

Interestingly, for \gls*{BiasedMNIST} we found that $\textit{learning rate} = 10^{-3}$ and $\textit{weight decay} = 10^{-5}$ worked best for all methods. Even though Up Wt and GDRO are known to generalize to minority groups when using a low learning rate and high weight decay~\cite{sagawa2020investigation}, we did not observe this for \gls*{BiasedMNIST}. We hypothesize that when  multiple sources of bias are present, as in \gls*{BiasedMNIST}, methods have multiple ways of predicting the classes, some of which maybe easier to learn than the others. When the hyperparameters are suitable to exploit these biases, methods obtain their best accuracies, which are still lower than \gls*{conventional}.

\section{Discussion}
\label{sec:discussions}

Our study demonstrates that systems are highly sensitive to the tuning distribution, that explicit methods cannot handle multiple bias sources, and that more rigorous analysis is critical for bias mitigation algorithms for future progress. Based on our results, we argue that the community should focus on implicit methods, rather than explicit, not only because explicit methods require additional annotations, but also because they perform worse on realistic settings. 

\paragraph{We make the following recommendations:}
\begin{enumerate}[noitemsep,leftmargin=*]
    \item Compare against multiple state-of-the-art methods under fair settings.
    \item Test on datasets that enable control over the number and degrees of biases, including realistic datasets.
    \item Analyze generalization to both explicit and implicit sources of bias.
    \item Be forthcoming about whether test distribution was used for model selection and compare robustness to tuning distributions that differ from the test.
\end{enumerate}
If these guidelines are adopted, we believe significant progress can be made so that bias mitigation algorithms can have real-world benefit for deployed systems.

\ifthenelse{\boolean{coordconv}}{
An interesting observation was that a weaker architecture, CNNs, were able to ignore position bias, whereas a more powerful architecture, CoordConv, resorted to exploiting this bias resulting in worse performance. While the community has largely focused on training procedures for bias mitigation, an exciting avenue for future work is to incorporate appropriate inductive biases into the architectures, perhaps endowing them with the ability to choose the the minimal computational power to do a task so that they are less sensitive to unwanted biases. This will essentially enable the algorithms to use Occam's razor to determine the minimal capabilities required to do a task to reduce their ability to utilize biases.
}

We have pointed to issues with the existing bias mitigation approaches, which alter the loss or use resampling. An orthogonal avenue for attacking bias mitigation is to use alternative architectures. Neuro-symbolic and graph-based systems could be created that focus on learning and grounding predictions on structured concepts, which have shown promising generalization capabilities~\cite{yi2018neural,mao2019neuro,hudson2019learning,gao2020multi,shi2019explainable}. Causality is another relevant line of research, where the goal is to uncover the underlying causal mechanisms~\cite{peters2016causal,meinshausen2018causality,bellot2020accounting,agarwal2020towards}. Discovery and usage of causal concepts is a promising direction for building robust systems. These areas have not been explicitly studied for their ability to overcome dataset bias.

\ifthenelse{\boolean{ack}}{
\paragraph{Acknowledgements.}
This work was supported in part by the DARPA/SRI Lifelong Learning Machines program [HR0011-18-C-0051], AFOSR grant [FA9550-18-1-0121], and NSF award \#1909696. The views and conclusions contained herein are those of the authors and should not be interpreted as representing the official policies or endorsements of any sponsor.
}

{\small
\bibliographystyle{ieee_fullname}
\bibliography{egbib}
}

\clearpage

\renewcommand{\thepage}{A\arabic{page}}  
\renewcommand{\thesection}{A}   
\renewcommand{\thetable}{A\arabic{table}}   
\renewcommand{\thefigure}{A\arabic{figure}}
\clearpage
\section{Appendix}

\begin{figure}[ht]
    \centering
    \includegraphics[width=0.45\textwidth]{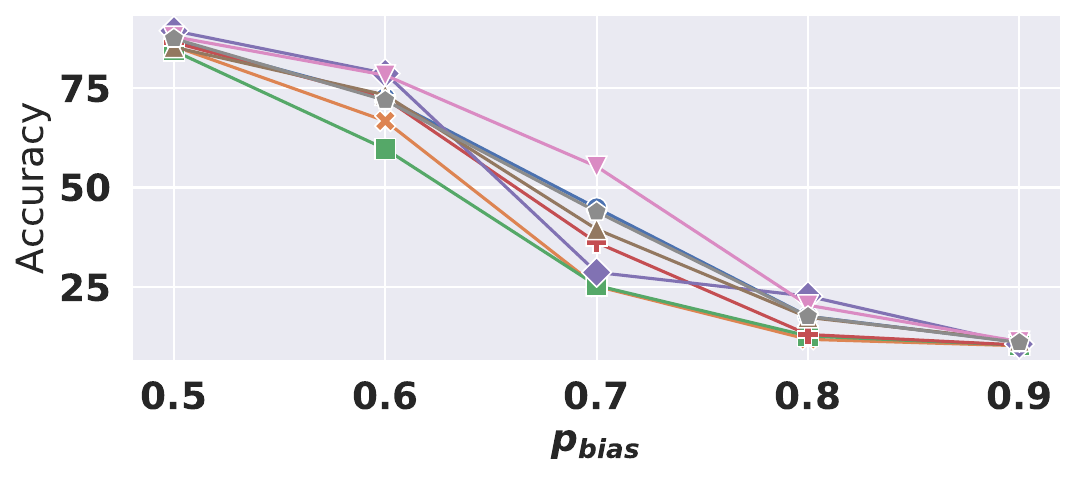}
    \includegraphics[width=0.35\textwidth]{images/legends/methods-legend-2.pdf}
    \caption{Unbiased accuracy on \gls*{BiasedMNIST} as a function $p_{bias}$.}
    \label{fig:biased_mnist_p_bias_overall}
\end{figure}

\begin{figure*}[]
    \centering
    \includegraphics[width=0.48\textwidth]{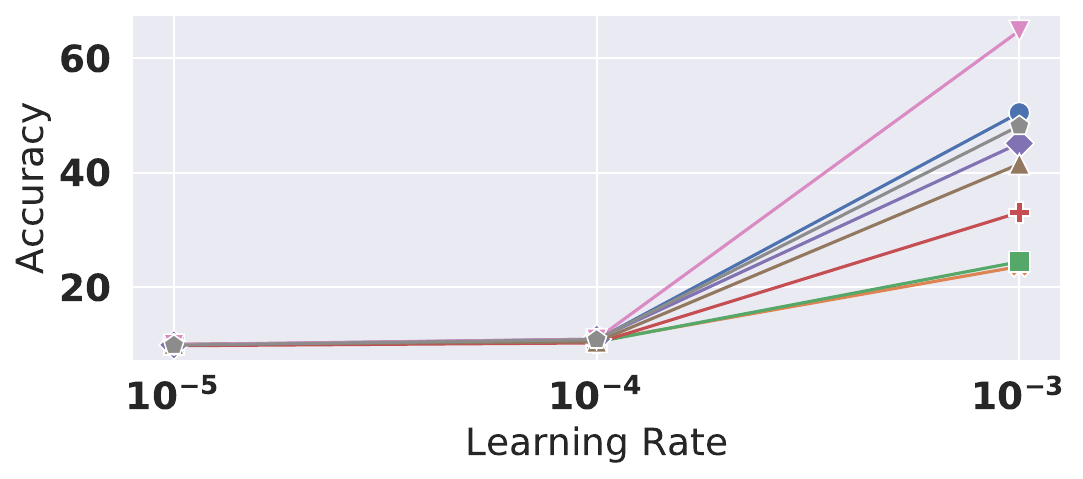}
    \includegraphics[width=0.48\textwidth]{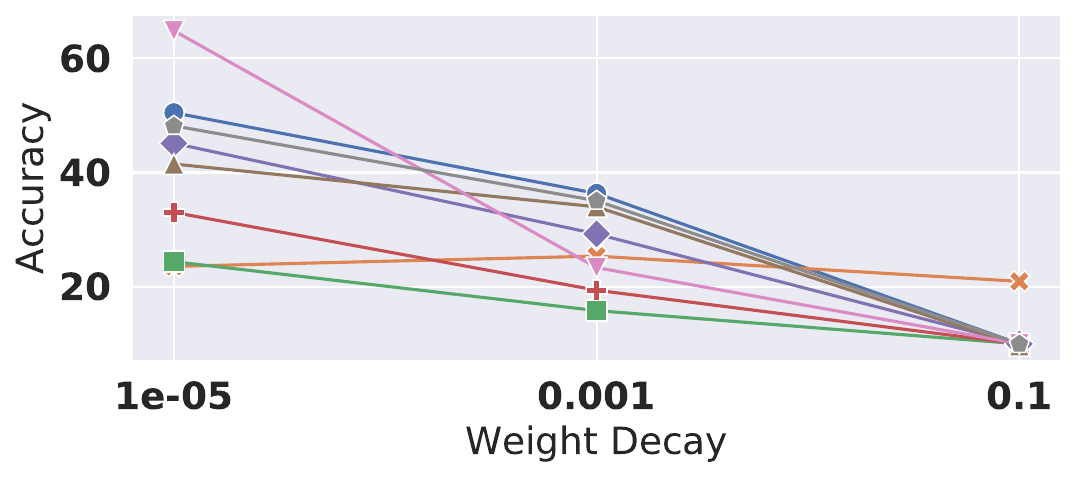}
    \includegraphics[width=0.9\textwidth]{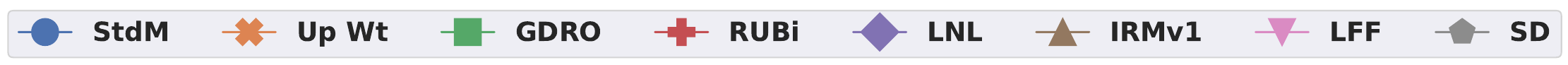}
    \caption{Unbiased test accuracy (mean per group) as a function of learning rate and weight decay (using single random seed).}
    \label{fig:biased_mnist_lr_wd}
\end{figure*}

\subsection{Additional Results}
We now provide additional results for all datasets, that provide further details for the results presented in Sec.~\ref{sec:analysis}.

\label{sec:appendix_biased_mnist_results}
\textbf{\gls*{BiasedMNIST}.} First, we show the unbiased accuracy as a function of $p_{bias}$ in Fig.~\ref{fig:biased_mnist_p_bias_overall}. None of the methods exhibit robustness across the range of $p_{bias}$ values, with all methods showing drops in the unbiased accuracy as $p_{bias}$ increases from 0.5 to 0.9. Second, in Fig.~\ref{fig:biased_mnist_lr_wd}, we show the unbiased accuracies as functions of learning rate and weight decay, when $p_{bias}$ is set to 0.7. The figure on the left shows unbiased accuracies as a function of learning rate with a fixed weight decay of $10^{-5}$ and the figure on the right shows unbiased accuracies as a function of weight decay, with learning rate set to $10^{-3}$. Barring Up Wt, all methods show highest unbiased accuracies at learning rate = $10^{-3}$ and weight decay = $10^{-5}$, and show large drops with other values, indicating high sensitivities to hyperparameters.

\begin{figure*}[t]
    \centering
    \includegraphics[width=0.98\textwidth]{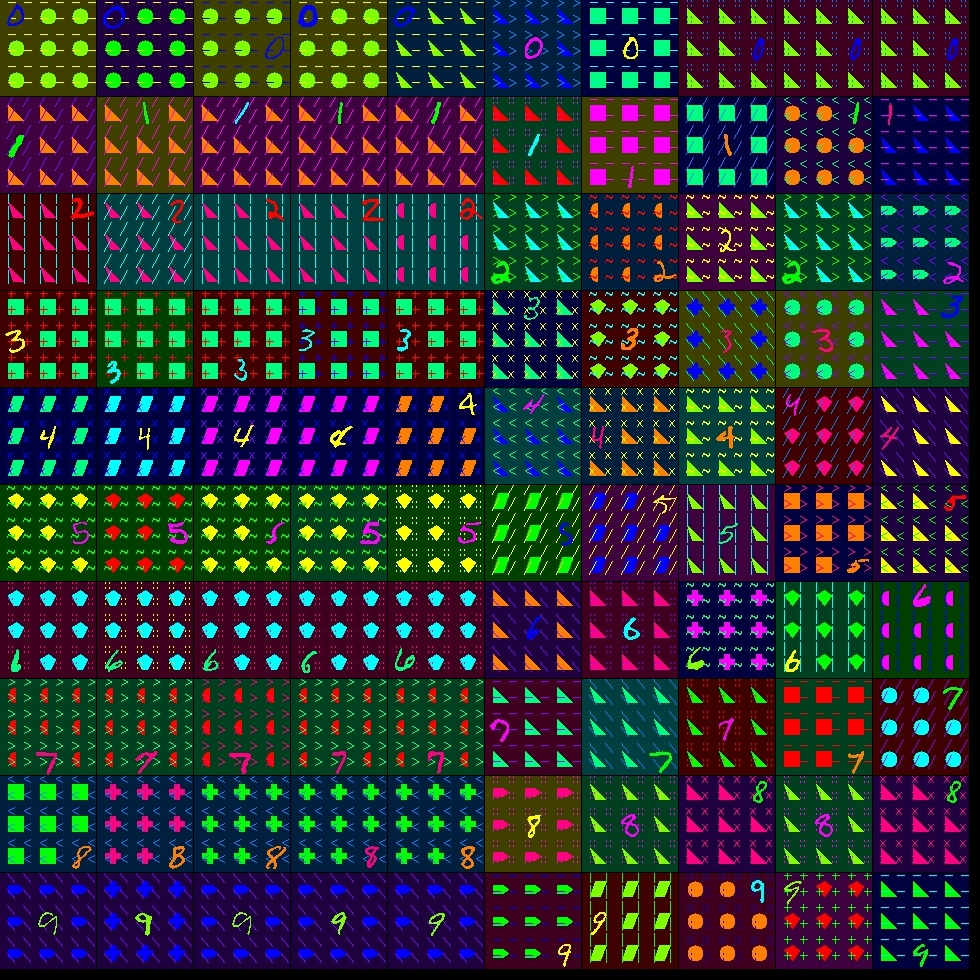}
    \caption{\gls*{BiasedMNIST} requires the methods to classify the target digit while remaining invariant to biases.}
    \label{fig:biased_mnist_composite_10x10}
\end{figure*}

Third, in Table.~\ref{tbl:biased_mnist_individual_expl_bias}, we compare the explicit methods with each other, when considering each of the seven variables as explicit bias variables ($b_{expl.}$) in separate experiments. For the explicit methods, variables apart from $b_{expl.}$ act as implicit biases, whereas implicit methods are not affected by the choice of the explicit variables. As previously discussed in Sec.~\ref{sec:bias_exploitation}, Up Wt, GDRO and RUBi show higher majority/minority difference for $b_{expl.}$ as compared to other variables, and all explicit methods are below \gls*{conventional} in terms of unbiased accuracies. Finally, in Table.~\ref{tbl:biased_mnist_multiple_expl_bias}, we show the per variable majority/minority accuracies when considering combinations of variables as explicit biases. As discussed in Sec.~\ref{sec:scalability_of_methods}, explicit methods barring Up Wt fail when multiple variables are specified as explicit bias variables. The results have been averaged across three runs for both tables. We provide a ten samples for each digit in Fig.~\ref{fig:biased_mnist_composite_10x10}.

\begin{table}[]
\centering
\caption{Per group accuracies on CelebA.}
\label{tbl:celebA_group_acc}
\begin{tabular}{lccccc} \hline
\multirow{2}{*}{\begin{tabular}[c]{@{}l@{}}Groups/\\ Methods\end{tabular}} &
  \multicolumn{2}{c}{Male} &
  \multicolumn{2}{c}{Non-Male} &
  \multirow{2}{*}{\begin{tabular}[c]{@{}c@{}}Unbiased\\ Accuracy\end{tabular}} \\
 &
  \begin{tabular}[c]{@{}c@{}}Non-\\ Blond\end{tabular} &
  Blond &
  \begin{tabular}[c]{@{}c@{}}Non-\\ Blond\end{tabular} &
  Blond &
   \\ \hline
 \rowcolor[HTML]{D0D0D0} 
\gls*{conventional}  & \textbf{\textit{\underline{99.3}}} & 42.8 & \textbf{\textit{\underline{95.9}}} & 83.3 & 80.3 \\
Up Wt & 89.3 & \textbf{77.8} & 87.4 & \textbf{95.0} & \underline{87.4} \\
GDRO  & 91.2 & \textbf{\textit{\underline{80.0}}} & 90.9 & \underline{92.1} & \textbf{88.5} \\
RUBi  & \underline{96.6} & 68.9 & \underline{94.3} & 89.1 & 87.2 \\
LNL   & \textbf{99.1} & 39.4 & \textbf{95.4} & 82.8 & 79.2 \\
IRMv1 & \textbf{99.1} & 40.6 & \textbf{\textit{\underline{95.9}}} & 83.6 & 79.8 \\
\rowcolor[HTML]{D0D0D0} 
LFF   & 96.0 & 46.1 & 90.4 & 80.0 & 77.8 \\
\rowcolor[HTML]{D0D0D0} 
SD    & 96.4 & \underline{73.3} & 88.0 & \textbf{\textit{\underline{96.6}}} & \textbf{\textit{\underline{88.6}}} \\ \hline
\end{tabular}%
\end{table}

\textbf{CelebA.} \label{sec:appendix_celebA_results} We show accuracy for each group of CelebA in Table.~\ref{tbl:celebA_group_acc}. SD and GDRO obtain the highest accuracies. As discussed previously, we observe trade-offs between blond and non-blond classes with the improvements in the rare blond class incurring degradations in the non-blond class.

\begin{figure*}[ht]
    \centering
    \subcaptionbox{Explicit Bias = Head vs Tail (\# groups = 2)\label{fig:gqa_tail_head_vs_tail}}{\includegraphics[width=0.46\textwidth]{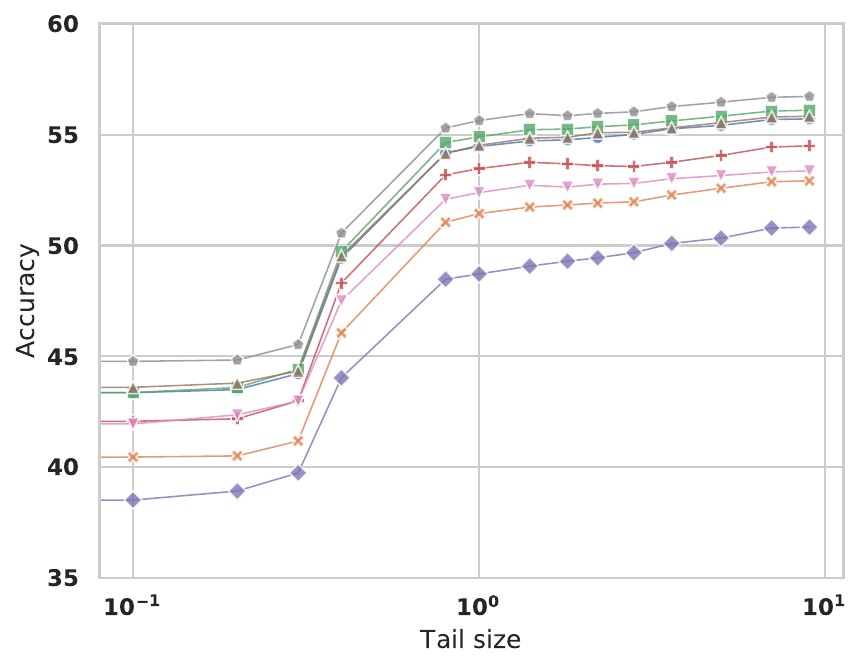}}
    \subcaptionbox{Explicit Bias = Global Group (\# groups = 115)\label{fig:gqa_tail_global_group}}{\includegraphics[width=0.46\textwidth]{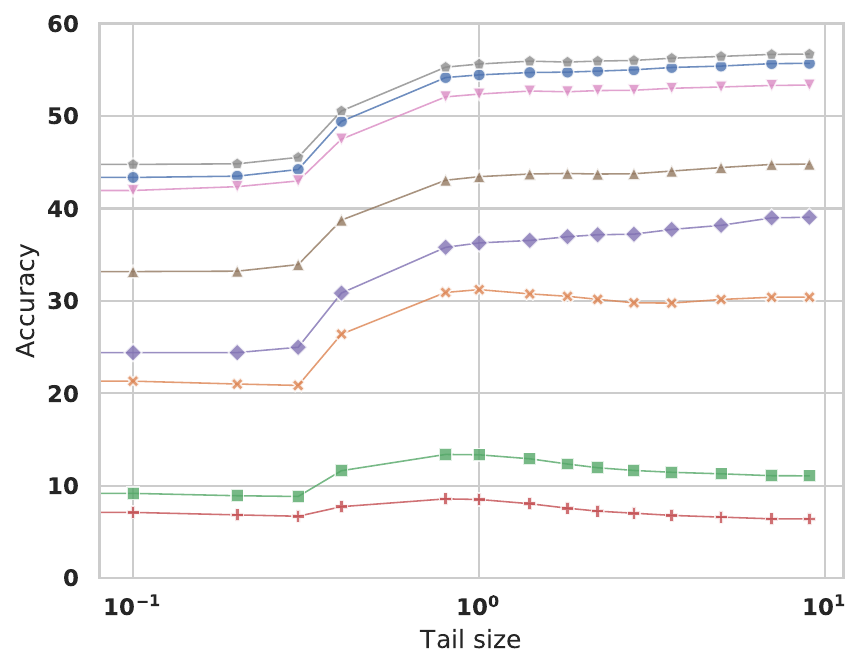}}
    \subcaptionbox{Explicit Bias = Answer Class (\# groups = 1833)\label{fig:gqa_tail_answer}}{\includegraphics[width=0.46\textwidth]{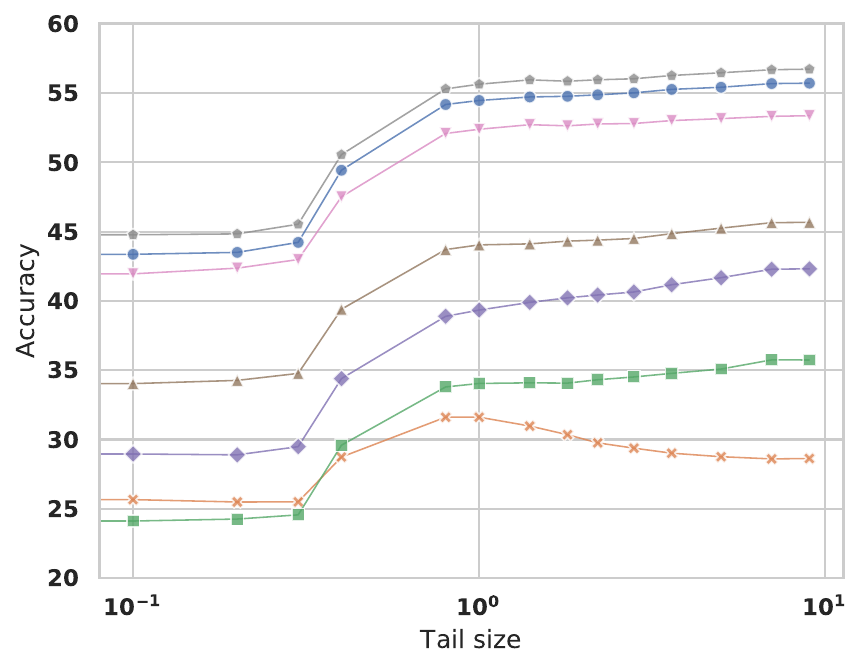}}
    \subcaptionbox{Explicit Bias = Local Group (\# groups = 133,328)\label{fig:gqa_tail_local_group}}{\includegraphics[width=0.46\textwidth]{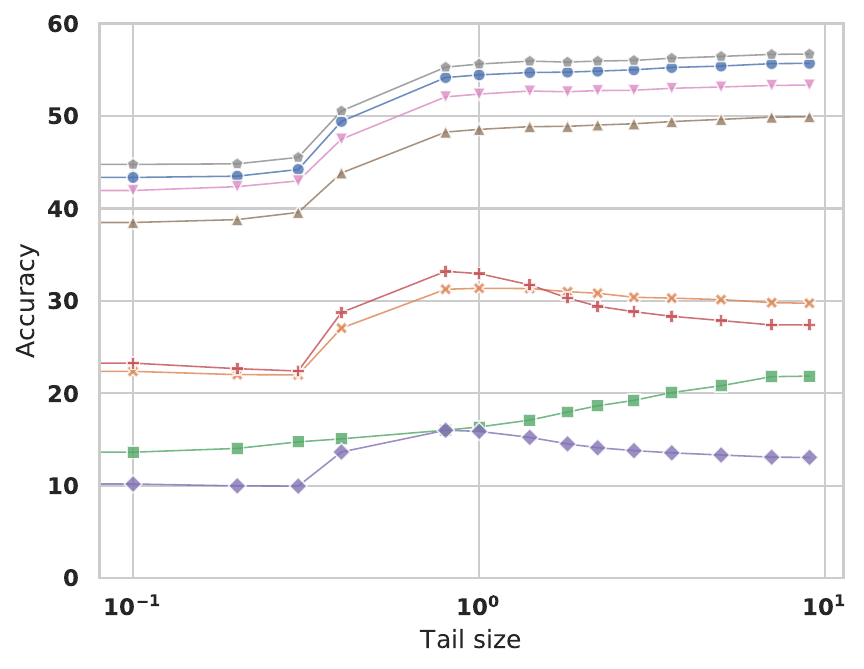}}
    \includegraphics[width=0.9\textwidth]{images/legends/methods-legend-1.pdf}
    \caption{Tail accuracies on GQA-OOD, when considering 4 different explicit biases.}
    \label{fig:gqa_tail}
\end{figure*}

\textbf{GQA-OOD.} GQA-OOD~\cite{kervadec2020roses} defines the tail accuracy (Acc-tail) metric which is computed on the samples of the tail of the answer class distribution. Specifically, an answer class $a_i$ is considered to be a tail class for a local group if:
\begin{align*}
    |a_i| \le (1 + \beta) \mu(a),
\end{align*}
where, $|a_i|$ is the number of instances for answer $a_i$ in the given group, $\mu(a)$ is the mean number of answers in the group and $\beta$ can be used to control the tail size. In Fig.~\ref{fig:gqa_tail}, we plot the tail accuracies at different tail sizes, considering different explicit bias variables for the explicit methods. For implicit methods: \gls*{conventional}, LFF and SD, same tail accuracies are repeated on all four charts since they are not affected by the choice of explicit variables during training. Explicit methods fail when the explicit variables entail generalization to large number of groups, whereas implicit methods are close to or above  \gls*{conventional}.

\subsection{Hyperparameters and Other Details}
We select hyperparameters based on the best unbiased validation set accuracy on each dataset, which is reflective of the unbiased test distribution. For all datasets and methods, we first perform a grid search over the learning rates $\in$ \{1e-3, 1e-4, 1e-5\} and weight decays $\in$ \{0, 0.1, 1e-3, 1e-5\}, and then tune the method-specific hyperparameters. For \gls*{BiasedMNIST}, the hyperparameters were selected using single run, considering `distractor shape' as the explicit bias variable for explicit methods. For CelebA, they were selected based on the unbiased accuracy/mean per group on the validation set and for GQA-OOD, they were selected based on the best mean head/tail accuracy when setting $\beta=0.2$ (the default value in the original paper). 

Next we specify the ranges considered for method-specific hyperparameters. For GDRO, we search the group weight step size between $\{0.001, 0.01, 0.1\}$. For LNL, we perform a grid search over gradient reversal weights $\lambda_{adv} \in \{-1.0, -0.1, -0.01\}$ and entropy loss weights $\lambda_{ent} \in \{1.0, 0.1, 0.01, 0\}$. For IRMv1, we search for $\lambda_{grad}$ values from \{1, 10, 100, 1000, 10000\}. The original implementation of IRMv1 samples from all environments in a single mini-batch during training. While this is feasible for small scale problems with few environments e.g., four environments/explicit groups for CelebA, it is computationally infeasible for \gls*{BiasedMNIST} and GQA, where the number of environments is larger than the batch size itself. So, for \gls*{BiasedMNIST} and GQA, we sample from 16 randomly selected groups or environments within each mini-batch during training. This implies that our implementation of IRMv1 samples uniformly from all environments.

Even though the original paper proposing LFF uses a fixed $\gamma=0.7$ for all datasets, we search over $\gamma \in \{0.1, 0.3, 0.5, 0.7, 0.9\}$ and find that LFF is indeed sensitive to $\gamma$. The default value: $\gamma=0.7$ is not the optimal value for all cases. For SD, we consider $\gamma,\lambda \in \{10^{-3}, 10^{-2}, 0.1, 1.0, 10.0, 100.0\}$. For CelebA, we use the class-specific $\gamma$ values specified in the paper and only tune $\lambda$. The procedure for obtaining class-specific hyperparameters when there are large number of classes e.g., in GQA however remains unclear for SD. The complete specification of hyperparameters is provided in Table.~\ref{tbl:hyperparameters}.

\begin{table*}[]
\centering
\caption{Majority (Maj.) and Minority (Min.) group accuracies for each variable in \gls*{BiasedMNIST} when using one of the seven variables as explicit bias for the explicit methods.}
\label{tbl:biased_mnist_individual_expl_bias}
\resizebox{\textwidth}{!}{%
\begin{tabular}{lccccccccccccccc} \hline
\multicolumn{1}{c}{\multirow{2}{*}{\begin{tabular}[c]{@{}c@{}}Variables/\\ Methods\end{tabular}}} &
  \multicolumn{1}{l}{} &
  \multicolumn{2}{c}{\begin{tabular}[c]{@{}c@{}}Background\\ Color\end{tabular}} &
  \multicolumn{2}{c}{\begin{tabular}[c]{@{}c@{}}Digit\\ Color\end{tabular}} &
  \multicolumn{2}{c}{\begin{tabular}[c]{@{}c@{}}Digit\\ Position\end{tabular}} &
  \multicolumn{2}{c}{\begin{tabular}[c]{@{}c@{}}Distractor\\ Shape\end{tabular}} &
  \multicolumn{2}{c}{\begin{tabular}[c]{@{}c@{}}Distractor \\ Color\end{tabular}} &
  \multicolumn{2}{c}{Texture} &
  \multicolumn{2}{c}{\begin{tabular}[c]{@{}c@{}}Texture \\ Color\end{tabular}} \\
\multicolumn{1}{c}{} &
  \begin{tabular}[c]{@{}c@{}}Unbiased \\ Accuracy\end{tabular} &
  \multicolumn{1}{l}{Maj.} &
  \multicolumn{1}{l}{Min.} &
  \multicolumn{1}{l}{Maj.} &
  \multicolumn{1}{l}{Min.} &
  \multicolumn{1}{l}{Maj.} &
  \multicolumn{1}{l}{Min.} &
  \multicolumn{1}{l}{Maj.} &
  \multicolumn{1}{l}{Min.} &
  \multicolumn{1}{l}{Maj.} &
  \multicolumn{1}{l}{Min.} &
  \multicolumn{1}{l}{Maj.} &
  \multicolumn{1}{l}{Min.} &
  \multicolumn{1}{l}{Maj.} &
  \multicolumn{1}{l}{Min.} \\ \hline
StdM  & 42.0 & 70.3 & 39.1 & 74.6 & 38.7 & 48.3 & 41.6 & 69.2 & 39.2 & 68.9 & 39.4 & 69.4 & 39.1 & 65.2 & 39.7 \\ \hline
\multicolumn{16}{c}{$b_{expl.}$ = Background Color}                                                            \\
Up Wt & 39.0 & 42.0 & 38.9 & 72.7 & 35.4 & 43.3 & 38.7 & 71.2 & 35.6 & 70.3 & 35.8 & 67.6 & 35.9 & 65.4 & 36.2 \\
GDRO  & 32.8 & 52.2 & 30.8 & 66.8 & 29.2 & 37.0 & 32.5 & 62.4 & 29.6 & 64.1 & 29.6 & 64.5 & 29.3 & 62.6 & 29.6 \\
RUBi  & 47.9 & 51.9 & 47.7 & 78.8 & 44.6 & 52.4 & 47.6 & 77.4 & 44.8 & 77.8 & 44.9 & 76.5 & 44.8 & 72.9 & 45.2 \\
LNL   & 44.0 & 55.8 & 43.5 & 59.7 & 43.0 & 46.8 & 44.5 & 57.5 & 43.3 & 57.6 & 43.3 & 56.6 & 43.3 & 53.2 & 43.7 \\
IRMv1 & 38.0 & 66.0 & 35.0 & 70.5 & 34.5 & 41.2 & 37.8 & 68.1 & 34.8 & 68.3 & 34.9 & 64.5 & 35.1 & 65.6 & 35.0 \\ \hline
\multicolumn{16}{c}{$b_{expl.}$ = Digit Color}                                                                 \\
Up Wt & 19.4 & 48.6 & 16.1 & 22.8 & 19.0 & 21.5 & 19.2 & 52.0 & 15.7 & 47.7 & 16.3 & 50.5 & 15.7 & 44.9 & 16.5 \\
GDRO  & 22.6 & 50.1 & 19.6 & 45.7 & 20.1 & 25.7 & 22.3 & 52.1 & 19.3 & 52.3 & 19.4 & 52.0 & 19.2 & 51.5 & 19.4 \\
RUBi  & 24.0 & 54.5 & 20.6 & 27.9 & 23.6 & 25.0 & 23.9 & 57.3 & 20.2 & 55.4 & 20.6 & 55.4 & 20.3 & 52.6 & 20.7 \\
LNL   & 40.2 & 53.7 & 39.5 & 57.9 & 39.0 & 46.2 & 40.3 & 56.1 & 39.2 & 55.1 & 39.4 & 56.1 & 39.2 & 50.9 & 39.8 \\
IRMv1 & 40.9 & 66.8 & 38.4 & 72.3 & 37.8 & 43.8 & 41.0 & 71.1 & 37.9 & 68.9 & 38.3 & 67.8 & 38.2 & 66.5 & 38.4 \\ \hline
\multicolumn{16}{c}{$b_{expl.}$ = Digit Position}                                                              \\
Up Wt & 21.4 & 46.7 & 18.6 & 50.4 & 18.2 & 21.9 & 21.4 & 53.4 & 17.8 & 50.4 & 18.3 & 49.0 & 18.2 & 45.8 & 18.6 \\
GDRO  & 26.8 & 54.9 & 23.9 & 55.4 & 23.9 & 27.6 & 27.0 & 58.2 & 23.5 & 54.9 & 24.0 & 54.5 & 23.9 & 51.6 & 24.3 \\
RUBi  & 26.6 & 54.9 & 23.5 & 59.1 & 23.0 & 28.3 & 26.5 & 56.8 & 23.2 & 56.9 & 23.4 & 56.0 & 23.2 & 53.5 & 23.6 \\
LNL   & 41.4 & 51.8 & 40.9 & 56.6 & 40.4 & 46.4 & 41.5 & 54.4 & 40.6 & 53.7 & 40.7 & 52.7 & 40.7 & 49.4 & 41.1 \\
IRMv1 & 38.1 & 67.8 & 35.0 & 70.8 & 34.7 & 42.9 & 37.8 & 69.3 & 34.8 & 68.0 & 35.1 & 67.0 & 34.9 & 66.0 & 35.1 \\ \hline
\multicolumn{16}{c}{$b_{expl.}$ = Distractor Shape}                                                                  \\
Up Wt & 21.5 & 52.0 & 18.2 & 56.0 & 17.8 & 24.2 & 21.4 & 25.6 & 21.2 & 53.0 & 18.2 & 54.0 & 17.9 & 51.9 & 18.2 \\
GDRO  & 25.2 & 54.1 & 22.2 & 56.9 & 21.9 & 27.9 & 25.2 & 48.9 & 22.8 & 56.0 & 22.1 & 56.0 & 21.9 & 52.3 & 22.4 \\
RUBi  & 34.5 & 66.9 & 31.0 & 70.9 & 30.5 & 35.2 & 34.5 & 40.4 & 33.9 & 67.7 & 31.0 & 67.5 & 30.7 & 64.6 & 31.1 \\
LNL   & 44.1 & 55.8 & 43.1 & 59.6 & 42.6 & 50.5 & 43.7 & 57.9 & 42.8 & 57.4 & 42.9 & 58.8 & 42.7 & 55.0 & 43.1 \\
IRMv1 & 38.3 & 67.0 & 35.5 & 70.5 & 35.1 & 41.5 & 38.3 & 69.3 & 35.2 & 65.9 & 35.7 & 67.9 & 35.2 & 64.6 & 35.7 \\ \hline
\multicolumn{16}{c}{$b_{expl.}$ = Distractor Color}                                                            \\
Up Wt & 40.9 & 70.1 & 38.0 & 70.8 & 38.0 & 46.6 & 40.7 & 73.0 & 37.7 & 44.8 & 40.9 & 73.0 & 37.6 & 68.5 & 38.2 \\
GDRO  & 28.1 & 58.6 & 24.8 & 57.4 & 24.9 & 30.4 & 28.0 & 62.1 & 24.4 & 49.1 & 25.9 & 59.1 & 24.6 & 55.8 & 25.1 \\
RUBi  & 49.5 & 77.3 & 46.6 & 80.1 & 46.3 & 54.5 & 49.1 & 78.3 & 46.4 & 53.8 & 49.2 & 77.6 & 46.4 & 76.4 & 46.6 \\
LNL   & 38.9 & 52.4 & 38.2 & 55.9 & 37.8 & 46.0 & 38.9 & 56.6 & 37.7 & 52.8 & 38.2 & 56.0 & 37.7 & 51.5 & 38.3 \\
IRMv1 & 39.7 & 66.1 & 36.8 & 72.3 & 36.1 & 45.1 & 39.2 & 68.8 & 36.5 & 69.1 & 36.6 & 66.4 & 36.6 & 64.8 & 36.9 \\ \hline
\multicolumn{16}{c}{$b_{expl.}$ = Texture}                                                                     \\
Up Wt & 29.3 & 60.8 & 25.9 & 66.1 & 25.4 & 32.9 & 29.1 & 64.6 & 25.5 & 62.3 & 25.9 & 32.9 & 29.1 & 60.7 & 25.9 \\
GDRO  & 24.5 & 54.9 & 21.3 & 58.0 & 21.0 & 27.3 & 24.4 & 56.8 & 21.1 & 54.8 & 21.4 & 42.7 & 22.6 & 51.6 & 21.6 \\
RUBi  & 43.1 & 73.7 & 39.8 & 78.5 & 39.3 & 44.9 & 43.1 & 74.2 & 39.7 & 74.8 & 39.8 & 48.1 & 42.7 & 71.7 & 40.0 \\
LNL   & 41.7 & 55.9 & 40.8 & 59.1 & 40.5 & 48.4 & 41.7 & 56.3 & 40.8 & 56.5 & 40.8 & 57.1 & 40.6 & 53.7 & 41.0 \\
IRMv1 & 37.5 & 64.4 & 34.6 & 67.1 & 34.3 & 39.5 & 37.4 & 68.0 & 34.2 & 65.8 & 34.5 & 66.2 & 34.3 & 62.5 & 34.8 \\ \hline
\multicolumn{16}{c}{$b_{expl.}$ = Texture Color}                                                               \\
Up Wt & 39.5 & 71.6 & 36.1 & 71.4 & 36.2 & 43.7 & 39.3 & 70.3 & 36.3 & 67.6 & 36.7 & 69.4 & 36.2 & 42.4 & 39.4 \\
GDRO  & 30.4 & 59.5 & 27.4 & 60.7 & 27.2 & 33.5 & 30.3 & 61.6 & 27.1 & 60.9 & 27.3 & 61.9 & 26.9 & 45.1 & 28.9 \\
RUBi  & 46.8 & 74.0 & 43.8 & 76.6 & 43.5 & 48.9 & 46.6 & 73.8 & 43.8 & 74.0 & 43.9 & 75.7 & 43.5 & 49.2 & 46.6 \\
LNL   & 34.3 & 49.2 & 33.4 & 47.9 & 33.5 & 41.8 & 34.2 & 49.9 & 33.3 & 47.8 & 33.6 & 48.6 & 33.4 & 43.3 & 34.0 \\
IRMv1 & 38.5 & 66.1 & 35.7 & 71.0 & 35.2 & 43.7 & 38.3 & 69.2 & 35.4 & 67.6 & 35.7 & 68.2 & 35.4 & 63.6 & 36.0 \\ \hline
\multicolumn{16}{c}{Implicit Methods}                                                                          \\
LFF   & 56.6 & 71.2 & 55.4 & 82.6 & 54.2 & 63.8 & 56.3 & 77.0 & 54.8 & 74.6 & 55.1 & 76.2 & 54.8 & 75.4 & 54.9 \\
SD    & 41.3 & 69.5 & 38.3 & 71.2 & 38.1 & 46.3 & 40.9 & 72.2 & 38.0 & 72.1 & 38.1 & 71.1 & 38.0 & 70.5 & 38.2 \\ \hline
\end{tabular}%
}
\end{table*}

\begin{table*}[]
\centering
\caption{Method accuracies with increasing number of explicit bias variables.}
\label{tbl:biased_mnist_multiple_expl_bias}
\resizebox{\textwidth}{!}{%
\begin{tabular}{cccccccccccccccc} \hline
\multicolumn{1}{l}{\multirow{2}{*}{\begin{tabular}[c]{@{}l@{}}Variables/\\ Methods\end{tabular}}} &
  \multicolumn{1}{l}{} &
  \multicolumn{2}{c}{\begin{tabular}[c]{@{}c@{}}Background\\ Color\end{tabular}} &
  \multicolumn{2}{c}{\begin{tabular}[c]{@{}c@{}}Digit \\ Color\end{tabular}} &
  \multicolumn{2}{c}{\begin{tabular}[c]{@{}c@{}}Digit \\ Position\end{tabular}} &
  \multicolumn{2}{c}{\begin{tabular}[c]{@{}c@{}}Distractor\\ Shape\end{tabular}} &
  \multicolumn{2}{c}{\begin{tabular}[c]{@{}c@{}}Distractor \\ Color\end{tabular}} &
  \multicolumn{2}{c}{\begin{tabular}[c]{@{}c@{}}Texture\\ Pattern\end{tabular}} &
  \multicolumn{2}{c}{\begin{tabular}[c]{@{}c@{}}Texture \\ Color\end{tabular}} \\
\multicolumn{1}{l}{} &
  \multicolumn{1}{l}{\begin{tabular}[c]{@{}l@{}}Unbiased \\ Accuracy\end{tabular}} &
  \multicolumn{1}{l}{Maj.} &
  \multicolumn{1}{l}{Min.} &
  \multicolumn{1}{l}{Maj.} &
  \multicolumn{1}{l}{Min.} &
  \multicolumn{1}{l}{Maj.} &
  \multicolumn{1}{l}{Min.} &
  \multicolumn{1}{l}{Maj.} &
  \multicolumn{1}{l}{Min.} &
  \multicolumn{1}{l}{Maj.} &
  \multicolumn{1}{l}{Min.} &
  \multicolumn{1}{l}{Maj.} &
  \multicolumn{1}{l}{Min.} &
  \multicolumn{1}{l}{Maj.} &
  \multicolumn{1}{l}{Min.} \\ \hline
StdM     & 42.0  & 70.3  & 39.1  & 74.6  & 38.7  & 48.3  & 41.6  & 69.2  & 39.2  & 68.9  & 39.4  & 69.4  & 39.1  & 65.2  & 39.7  \\ \hline
\multicolumn{16}{c}{$b_{expl.}$ = Distractor Shape}                                                                             \\
Up Wt    & 21.5  & 52.0  & 18.2  & 56.0  & 17.8  & 24.2  & 21.4  & 25.6  & 21.2  & 53.0  & 18.2  & 54.0  & 17.9  & 51.9  & 18.2  \\
GDRO     & 25.2  & 54.1  & 22.2  & 56.9  & 21.9  & 27.9  & 25.2  & 48.9  & 22.8  & 56.0  & 22.1  & 56.0  & 21.9  & 52.3  & 22.4  \\
RUBi     & 34.5  & 66.9  & 31.0  & 70.9  & 30.5  & 35.2  & 34.5  & 40.4  & 33.9  & 67.7  & 31.0  & 67.5  & 30.7  & 64.6  & 31.1  \\
LNL      & 44.1  & 55.8  & 43.1  & 59.6  & 42.6  & 50.5  & 43.7  & 57.9  & 42.8  & 57.4  & 42.9  & 58.8  & 42.7  & 55.0  & 43.1  \\
IRMv1    & 38.3  & 67.0  & 35.5  & 70.5  & 35.1  & 41.5  & 38.3  & 69.3  & 35.2  & 65.9  & 35.7  & 67.9  & 35.2  & 64.6  & 35.7  \\ \hline
\multicolumn{16}{c}{$b_{expl.}$ = Distractor Shape+Texture Pattern}                                                             \\
Up Wt    & 16.1  & 49.0  & 12.6  & 51.6  & 12.3  & 18.3  & 16.0  & 20.6  & 15.8  & 50.7  & 12.5  & 18.5  & 16.0  & 45.3  & 12.9  \\
GDRO     & 17.4  & 43.7  & 14.6  & 44.3  & 14.6  & 20.1  & 17.3  & 42.5  & 14.8  & 46.4  & 14.4  & 42.6  & 14.6  & 41.8  & 14.8  \\
RUBi     & 27.9  & 64.3  & 24.0  & 70.9  & 23.3  & 32.1  & 27.6  & 36.5  & 27.1  & 65.9  & 24.0  & 34.1  & 27.4  & 60.6  & 24.3  \\
LNL      & 10.0  & 14.7  & 10.5  & 11.0  & 10.9  & 11.3  & 10.9  & 12.3  & 10.8  & 11.8  & 10.9  & 12.9  & 10.7  & 11.7  & 10.9  \\
IRMv1    & 34.3  & 64.4  & 31.2  & 68.5  & 30.8  & 38.5  & 34.2  & 63.9  & 31.3  & 63.7  & 31.4  & 61.3  & 31.4  & 62.3  & 31.4  \\ \hline
\multicolumn{16}{c}{$b_{expl.}$ = Distractor Shape+Texture Pattern+Digit Color}                                                 \\
Up Wt    & 12.6  & 47.7  & 8.7   & 20.1  & 11.8  & 13.5  & 12.5  & 22.9  & 11.5  & 45.8  & 9.0   & 22.7  & 11.4  & 42.3  & 9.2   \\
GDRO     & 18.4  & 44.1  & 15.6  & 42.6  & 15.8  & 20.0  & 18.3  & 44.8  & 15.5  & 45.7  & 15.6  & 44.2  & 15.5  & 43.6  & 15.7  \\
RUBi     & 11.0  & 48.9  & 6.8   & 12.9  & 10.9  & 11.8  & 11.0  & 11.1  & 11.1  & 46.8  & 7.2   & 11.0  & 11.1  & 41.0  & 7.6   \\
LNL      & 10.2  & 15.2  & 10.6  & 11.5  & 11.1  & 11.7  & 11.0  & 11.4  & 11.1  & 12.2  & 11.0  & 18.4  & 10.2  & 15.3  & 10.6  \\
IRMv1    & 24.9  & 57.0  & 21.4  & 53.8  & 21.8  & 27.6  & 24.8  & 52.4  & 21.9  & 56.0  & 21.7  & 50.7  & 22.0  & 55.3  & 21.6  \\ \hline
\multicolumn{16}{c}{$b_{expl.}$ = Distractor Shape+Texture Pattern+Digit Color+Background Color}                                \\
Up Wt    & 30.1  & 54.5  & 27.6  & 59.3  & 27.1  & 35.0  & 29.8  & 56.0  & 27.4  & 64.0  & 26.7  & 56.3  & 27.3  & 63.4  & 26.5  \\
GDRO     & 19.9  & 48.2  & 17.0  & 45.3  & 17.4  & 22.8  & 19.9  & 50.9  & 16.7  & 47.4  & 17.2  & 46.7  & 17.1  & 43.8  & 17.5  \\
RUBi     & 10.0  & 10.3  & 9.9   & 11.9  & 9.7   & 10.0  & 10.0  & 9.2   & 10.0  & 59.4  & 4.6   & 10.4  & 9.9   & 48.0  & 5.6   \\
LNL      & 10.0  & 10.0  & 11.5  & 10.6  & 11.4  & 11.2  & 11.4  & 11.9  & 11.3  & 13.0  & 11.2  & 11.3  & 11.4  & 10.6  & 11.4  \\
IRMv1    & 25.6  & 52.9  & 22.7  & 58.1  & 22.1  & 30.1  & 25.2  & 52.3  & 22.7  & 57.5  & 22.3  & 52.4  & 22.6  & 56.0  & 22.3  \\ \hline
\multicolumn{16}{c}{$b_{expl.}$ = Distractor Shape+Texture Pattern+Digit Color+Background Color+Texture Color}                  \\
Up Wt    & 38.8  & 65.3  & 36.0  & 71.4  & 35.3  & 42.3  & 38.5  & 68.2  & 35.6  & 71.0  & 35.4  & 67.1  & 35.6  & 61.5  & 36.3  \\
GDRO     & 21.9  & 49.9  & 19.1  & 47.2  & 19.4  & 25.1  & 21.8  & 53.2  & 18.7  & 50.8  & 19.1  & 51.6  & 18.7  & 45.6  & 19.5  \\
RUBi     & 9.8   & 10.2  & 9.7   & 9.6   & 9.8   & 9.8   & 9.7   & 8.8   & 9.8   & 88.2  & 1.2   & 9.7   & 9.7   & 9.3   & 9.8   \\
LNL      & 10.0  & 10.0  & 11.5  & 10.6  & 11.4  & 11.2  & 11.4  & 11.9  & 11.3  & 13.0  & 11.2  & 11.3  & 11.4  & 10.6  & 11.4  \\
IRMv1    & 28.4  & 58.7  & 25.0  & 61.5  & 24.7  & 32.2  & 28.0  & 56.6  & 25.2  & 59.3  & 25.1  & 57.8  & 25.0  & 53.5  & 25.6  \\ \hline
\multicolumn{16}{c}{$b_{expl.}$ = Distractor Shape+Texture Pattern+Digit Color+Background Color+Texture Color+Distractor Color} \\
Up Wt    & 43.2  & 71.0  & 40.4  & 75.3  & 39.9  & 47.3  & 43.1  & 72.1  & 40.2  & 69.7  & 40.6  & 72.8  & 40.0  & 68.5  & 40.6  \\
GDRO     & 19.7  & 47.3  & 16.7  & 42.2  & 17.3  & 22.2  & 19.5  & 50.7  & 16.3  & 48.5  & 16.7  & 48.1  & 16.5  & 44.4  & 17.0  \\
RUBi     & 9.9   & 9.0   & 10.1  & 9.8   & 10.0  & 11.4  & 9.8   & 10.2  & 10.0  & 12.3  & 9.7   & 10.9  & 9.9   & 9.2   & 10.1  \\
LNL      & 10.4  & 22.4  & 9.6   & 12.1  & 10.8  & 13.5  & 10.6  & 11.3  & 10.9  & 11.5  & 10.8  & 30.8  & 8.6   & 24.8  & 9.3   \\
IRMv1    & 24.8  & 52.6  & 21.8  & 58.3  & 21.2  & 26.0  & 24.8  & 57.4  & 21.3  & 53.2  & 21.9  & 53.3  & 21.6  & 49.5  & 22.1  \\ \hline
\multicolumn{16}{c}{\begin{tabular}[c]{@{}c@{}}$b_{expl.}$ = Distractor Shape+Texture Pattern+Digit Color+Background Color+Texture Color+Distractor Color\\ +Digit Position\end{tabular}} \\
Up Wt    & 44.0  & 70.5  & 41.4  & 73.9  & 41.0  & 48.8  & 43.9  & 71.8  & 41.2  & 71.8  & 41.4  & 71.0  & 41.2  & 68.2  & 41.6  \\
GDRO     & 16.6  & 42.6  & 13.7  & 34.6  & 14.7  & 19.0  & 16.4  & 46.8  & 13.3  & 46.5  & 13.4  & 44.1  & 13.4  & 38.9  & 14.1  \\
RUBi     & 10.0  & 11.9  & 10.6  & 10.2  & 10.8  & 10.7  & 10.7  & 10.2  & 10.8  & 12.6  & 10.5  & 11.2  & 10.7  & 9.7   & 10.8  \\
LNL      & 10.0  & 10.1  & 11.5  & 10.5  & 11.4  & 11.1  & 11.4  & 12.2  & 11.2  & 12.8  & 11.2  & 11.4  & 11.3  & 10.7  & 11.4  \\
IRMv1    & 21.8  & 47.7  & 19.0  & 51.6  & 18.6  & 24.6  & 21.6  & 49.8  & 18.7  & 52.2  & 18.6  & 49.3  & 18.7  & 46.5  & 19.1  \\ \hline
\multicolumn{16}{c}{Implicit Methods}                                                                                            \\
LFF      & 56.6  & 71.2  & 55.4  & 82.6  & 54.2  & 63.8  & 56.3  & 77.0  & 54.8  & 74.6  & 55.1  & 76.2  & 54.8  & 75.4  & 54.9  \\
SD       & 41.3  & 69.5  & 38.3  & 71.2  & 38.1  & 46.3  & 40.9  & 72.2  & 38.0  & 72.1  & 38.1  & 71.1  & 38.0  & 70.5  & 38.2 \\ \hline
\end{tabular}%
}
\end{table*}

\begin{table*}[]
\centering
\caption{Hyperparameters for all datasets for all the comparison methods selected using unbiased accuracies of the validation sets after performing grid search over hyperparameters.}
\label{tbl:hyperparameters}
\resizebox{0.9\textwidth}{!}{%
\begin{tabular}{llccc}
\hline
\textbf{Methods/Datasets} &
  \textbf{Parameters} &
  \textbf{CelebA} &
  \textbf{Biased MNIST} &
  \textbf{GQA-OOD} \\ \hline
\multirow{3}{*}{\shortstack[1]{Common to all methods \\  unless specified otherwise}}  &
  Optimizer &
  SGD &
  Adam &
  Adam \\ 
 &
  Batch Size &
  128 &
  128 &
  128 \\
&
Epochs &
  50 &
  30 &
  30 \\ \hline
\multirow{2}{*}{\begin{tabular}[c]{@{}l@{}}Standard Model\\ (\gls*{conventional}) \end{tabular}} &
  Learning Rate &
  $10^{-3}$ &
  $10^{-3}$ &
  $10^{-4}$ \\ 
 &
  Weight Decay &
  0 &
  $10^{-5}$ &
  0 \\ \hline
\multirow{2}{*}{\begin{tabular}[c]{@{}l@{}}Group Upweighting\\ (Up Wt)~\cite{sagawa2020investigation} \end{tabular}} &
  Learning Rate &
  $10^{-5}$ &
  $10^{-3}$ &
  $10^{-3}$ \\ 
 &
  Weight Decay &
  0.1 &
  $10^{-5}$ &
  0 \\ \hline
\multirow{3}{*}{\begin{tabular}[c]{@{}l@{}}Group DRO\\ (GDRO)~\cite{sagawa2019distributionally} \end{tabular}} &
  Learning Rate &
  $10^{-5}$ &
  $10^{-3}$ &
  $10^{-4}$ \\ 
 &
  Weight Decay &
  0.1 &
  $10^{-5}$ &
  0 \\ 
 &
  Step Size &
  0.01 &
  $10^{-3}$ &
  0.01 \\ \hline
\multirow{2}{*}{\begin{tabular}[c]{@{}l@{}}Reduction of Unimodal \\ Biases (RUBi)~\cite{cadene2019rubi}\end{tabular}} &
  Learning Rate &
  $10^{-4}$ &
  $10^{-3}$ &
  $10^{-4}$ \\
 &
  Weight Decay &
  $10^{-5}$ &
  $10^{-5}$ &
  0 \\ \hline
\multirow{4}{*}{\begin{tabular}[c]{@{}l@{}}Adversarial Regularization\\ (LNL)~\cite{kim2019learning}\end{tabular}} &
  Learning Rate &
  $10^{-4}$ &
  $10^{-3}$ &
  $10^{-3}$ \\ 
 &
  Weight Decay &
  $10^{-4}$ &
  $10^{-5}$ &
  0 \\ 
 &
  Gradient Reversal Weight ($\lambda_{grad.}$) &
  -0.1 &
  -0.1 &
  -0.1 \\ 
 &
  Entropy Loss Weight ($\lambda_{ent.}$) &
  0 &
  0.01 &
  0.01 \\ \hline
\multirow{3}{*}{\begin{tabular}[c]{@{}l@{}}Invariant Risk Minimization\\ (IRMv1)~\cite{arjovsky2019invariant}\end{tabular}} &
  Learning Rate &
  $10^{-4}$ &
  $10^{-3}$ &
  $10^{-4}$ \\ 
 &
  Weight Decay &
  0 &
  $10^{-5}$ & 
  0 \\ 
 &
  Gradient Regularization Weight ($\lambda_{grad.}$) &
  1.0 &
  0.01 &
  0.01 \\
  &
Number of environments per mini-batch  &
  4 &
  16 &
  16 \\\hline  
\multirow{3}{*}{\begin{tabular}[c]{@{}l@{}}Learning From Failure\\ (LFF)~\cite{nam2020learning}\end{tabular}} &
  Optimizer & Adam & Adam & Adam \\
  & Learning Rate &
  $10^{-4}$ &
  $10^{-3}$ &
  $10^{-4}$ \\
 &
  Weight Decay &
  0 &
  $10^{-5}$ &
  0 \\ 
 &
  Amplification Factor ($\gamma$) &
  0.1 &
  0.5 &
  0.7 \\ \hline
\multirow{4}{*}{Spectral Decoupling (SD)~\cite{pezeshki2020gradient}} &
  Learning Rate &
  $10^{-4}$ &
  $10^{-3}$ &
  $10^{-4}$ \\ 
 &
  Weight Decay &
  $10^{-5}$ &
  $10^{-5}$ &
  0 \\ 
 &
  Output Decay ($\lambda$) &
  \begin{tabular}[c]{@{}c@{}}$\lambda_0$ = 10\\ $\lambda_1$ = 10\end{tabular} &
  $10^{-3}$ &
  $10^{-3}$ \\
 &
  Output Shift ($\gamma$) &
  \begin{tabular}[c]{@{}c@{}}$\gamma_0$ = 0.44\\ $\gamma_1$ = 2.5\end{tabular} &
  $10^{-3}$ &
  $10^{-3}$ \\ \hline
\end{tabular}%
}
\end{table*}

\end{document}